\documentclass[twoside,11pt]{article}
\usepackage{jair, theapa, rawfonts}

\usepackage{times}
\usepackage{helvet}
\usepackage{courier}
\usepackage{tikz}
\usetikzlibrary{arrows}
\usepackage{subcaption}
\usepackage{booktabs}
\usepackage{amsmath}
\usepackage{amssymb}
\usepackage{bm}
\usepackage{algorithm}
\usepackage{algorithmic}
\usetikzlibrary{positioning,calc}
\usetikzlibrary{fit}
\usepackage{varwidth}
\usepackage{enumitem}
\usepackage{color,soul}

\usepackage[font=small]{caption}

\usepackage{verbatim}
\usepackage{etoolbox}

\makeatletter
\patchcmd{\verbatim@input}{\@verbatim}{\scriptsize\@verbatim}{}{}
\makeatother

\newcommand{\actionset}{A}
\newcommand{\stateset}{S}
\newcommand{\initialstate}{I}
\newcommand{\goalstate}{G}
\newcommand{\rewardfunction}{Q} 
\newcommand{\transitionfunction}{T}
\newcommand{\actionstateconstraints}{C}
\newcommand{\horizon}{H}

\newcommand{\vnorm}[1]{\left\lVert#1\right\rVert}
\newcommand{\myvec}[1]{\boldsymbol{#1}}

\begin{document}

\title{Scalable Planning with Deep Neural Network \\Learned Transition Models}

\author{\name Ga Wu
\email wuga@mie.utoronto.ca \\
       \addr Department of Mechanical and Industrial Engineering\\
       University of Toronto, Toronto, ON, Canada\\
       Vector Institute for Artificial Intelligence, Toronto, ON, Canada\\
       \AND
       \name Buser Say \email buser.say@monash.edu \\
       \addr Department of Mechanical and Industrial Engineering\\
       University of Toronto, Toronto, ON, Canada\\
       Vector Institute for Artificial Intelligence, Toronto, ON, Canada\\
       Faculty of Information Technology\\
       Monash University, Melbourne, VIC, Australia\\
       \AND
       \name Scott Sanner \email ssanner@mie.utoronto.ca \\
       \addr Department of Mechanical and Industrial Engineering\\
       University of Toronto, Toronto, ON, Canada\\
       Vector Institute for Artificial Intelligence, Toronto, ON, Canada
}


\maketitle

\begin{abstract}
In many complex planning problems with factored, continuous state and action spaces such as Reservoir Control, Heating Ventilation and Air Conditioning 
(HVAC), and Navigation domains, it is difficult to obtain a model of the complex nonlinear dynamics that govern state evolution.  
However, the ubiquity of modern sensors allows us to collect large quantities of data from each of these complex systems and build accurate, nonlinear deep neural network models of their state transitions.  But there remains one major problem 
for the task of control -- how can we plan with deep network learned transition 
models without resorting to Monte Carlo Tree Search and other black-box transition model techniques that ignore model structure and do not easily extend to continuous domains? In this paper, we introduce two types of planning methods that can leverage deep neural network learned transition models: Hybrid Deep MILP Planner (HD-MILP-Plan) and Tensorflow Planner (TF-Plan). In HD-MILP-Plan, we make the critical observation that the Rectified Linear Unit (ReLU) transfer function for 
deep networks not only allows faster convergence of model learning, but also permits a direct compilation of the deep network transition model to a 
Mixed-Integer Linear Program (MILP) encoding. Further, we identify deep network specific optimizations for HD-MILP-Plan that improve performance over a base encoding and show that we can plan optimally with respect to the learned deep networks. In TF-Plan, we take advantage of the efficiency of auto-differentiation tools and GPU-based computation where we encode a subclass of purely continuous planning problems as Recurrent Neural Networks and directly optimize the actions through backpropagation. We compare both planners and show that TF-Plan is able to approximate the optimal plans found by HD-MILP-Plan in less computation time.  Hence this article offers two novel planners for continuous state and action domains with learned deep neural net transition models: one optimal method (HD-MILP-Plan) and a scalable alternative for large-scale problems (TF-Plan).
\end{abstract}

\section{Introduction}
\label{sec:intro}

In many complex planning problems with factored~\shortcite{Boutilier1999} and continuous state and action spaces such as Reservoir Control~\cite{Yeh1985}, Heating, Ventilation and Air Conditioning 
(HVAC)~\shortcite{agarwal2010}, and Navigation~\cite{nonlinear_path_planning}, 
it is difficult to obtain a model of the complex nonlinear dynamics that govern state 
evolution.  For example, in Reservoir Control, evaporation and other sources of water 
loss are a complex function of volume, bathymetry, and environmental conditions; in 
HVAC domains, thermal conductance between walls and convection properties of rooms 
are nearly impossible to derive from architectural layouts; and in Navigation problems, 
nonlinear interactions between surfaces and traction devices make it hard to accurately 
predict odometry.

A natural answer to these modeling difficulties is to instead learn the transition 
model from sampled data; fortunately, the presence of vast sensor networks often 
make such data inexpensive and abundant.  While learning nonlinear models with 
{\it a priori} unknown model structure can be very difficult in practice, recent 
progress in Deep Learning and the availability of off-the-shelf tools such as 
Tensorflow~\shortcite{abadi2016tensorflow} and Pytorch~\shortcite{paszke2017automatic} make it possible to learn highly accurate 
nonlinear deep neural networks with little prior knowledge of model structure.  

However, the modeling of a nonlinear transition model as a deep neural network poses 
non-trivial difficulties for the optimal control task.  Existing planners with nonlinear transition structure are  
either are not compatible with nonlinear deep network transition models and continuous 
(i.e., real-valued) actions\footnote{\shortcite{Penna2009,lohr,coles2013hybrid,ivankovic,piotrowski,scala2016interval}}, or 
only optimize goal-oriented objectives\footnote{\shortcite{Bryce2015,Scala2016-2,Cashmore2016}}. 
Monte Carlo Tree Search (MCTS) methods~\cite{mcts,uct,keller_icaps13} including 
AlphaGo~\shortcite{silver2016mastering} that \emph{could} exploit a deep network learned black box model 
of transition dynamics do not inherently work with continuous action spaces due to the 
infinite branching factor. While MCTS with continuous action extensions such as 
HOOT~\cite{weinstein2012bandit} have been proposed, their continuous partitioning 
methods do not scale to high-dimensional concurrent and continuous action spaces. 
Finally, offline model-free reinforcement 
learning with function approximation~\cite{sutton_barto,csaba_rl} and 
deep extensions~\shortcite{deepqn} do not directly apply to domains with high-dimensional 
continuous action spaces. That is, offline learning methods like Q-learning require 
action maximization for every update, but in high-dimensional continuous action 
spaces such nonlinear function maximization is non-convex and computationally 
intractable at the scale of millions or billions of updates. 

Despite these limitations of existing methods, all is not lost. First, we remark 
that our deep network is not a black-box but rather a gray-box; while 
the learned parameters often lack human interpretability, there is still a uniform 
layered symbolic structure in the deep neural network models.  Second, we make the critical observation 
that the popular Rectified Linear Unit (ReLU)~\cite{relu} transfer function for 
deep networks enables effective \emph{nonlinear} deep neural network model learning and permits 
a direct compilation to a Mixed-Integer Linear Program (MILP) encoding. Given other 
components such as a human-specified objective function and a horizon, this permits 
direct optimization in a method we call Hybrid Deep MILP Planner (HD-MILP-Plan).

While arguably an important step forward, we remark that planners with optimality guarantees such as  
HD-MILP-Plan can only scale up to moderate-sized planning problems.  Hence in an effort to scale to substantially larger control problems, we focus on a general subclass of planning problems with purely continuous state and action spaces in order to take advantage 
of the efficiency of auto-differentiation tools and GPU-based computation.  Specifically, we propose to extend work using the Tensorflow tool for planning~\shortcite{Wu2017} in deterministic continuous RDDL~\cite{Sanner:RDDL} domains to the case of learned neural network transition models investigated in this article.  Specifically, we show that we can embed both a  
reward function and a deep-learned transition function into a Recurrent Neural Network (RNN) cell, chain multiple of these RNN cells together for a fixed horizon, and produce a plan in the resulting RNN encoding through end-to-end backpropagation in a method we call Tensorflow Planner (TF-Plan).

In brief, we can summarize the high-level procedure of the two approaches we contribute in this article by the following two steps:
\begin{enumerate}
    \item Train a neural network to learn a transition model from sample trajectory data that predicts the next state given the current state and action. 
    \item Chain these (fixed) learned transition models together for a fixed planning horizon and optimize the action choices given a fixed initial state and an overall planning objective.  We contribute and evaluate two different encodings and optimizers for this task: 
    \begin{itemize} 
    \item HD-MILP-Plan, which compiles the learned ReLU-based neural network transition model and overall objective into a MILP and leverage the CPLEX optimizer\footnote{\texttt{http://www.cplex.com/}}.
    \item TF-Plan, which encodes the learned transition model in an RNN-based deep neural network unrolled for a fixed planning horizon and leverages Tensorflow~\cite{abadi2016tensorflow}\footnote{\texttt{https://www.tensorflow.org/}} for fast and efficient gradient descent through backpropagation to optimize the overall objective w.r.t. this (fixed) learned transition model.
    \end{itemize} 
\end{enumerate}


Experimentally, 
we compare HD-MILP-Plan and TF-Plan versus manually specified domain-specific policies on Reservoir Control, HVAC, and Navigation domains.  Our primary objectives are to comparatively evaluate the ability of HD-MILP-Plan and TF-Plan to produce high quality plans with limited computational resources in an online planning setting, and to assess their performance against carefully designed manual policies.  For HD-MILP-Plan, we show that our strengthened MILP encoding improves the quality of plans produced in less computational time over the base encoding. For TF-Plan, we show the scalability and the efficiency of our planner on large-scale problems and its ability to approximate the optimal plans found by HD-MILP-Plan on moderate-sized problems.  Overall, this article contributes and evaluates two novel approaches for planning in  continuous state and action domains with learned deep neural net transition models: one optimal method (HD-MILP-Plan) and a scalable alternative for large-scale problems (TF-Plan).

\section{Deterministic Factored Planning Problem Specification}
\label{sec:det_fac_planning}

Before we proceed to discuss deep network transition learning, we review the general planning problem that motivates this work.
A deterministic factored planning problem is a tuple $\Pi = \langle \stateset, \actionset, 
\actionstateconstraints, \transitionfunction, \initialstate, \goalstate, \rewardfunction 
\rangle$ 
where $\stateset = \{s^1,\dots, s^{n_1}\}$ and $\actionset  = \{a^1,\dots, a^{n_2}\}$ are sets of 
state and action variables with 
continuous domains\footnote{In this article we focus on purely continuous state and action planning problems.  While the extension to mixed (i.e., continuous and discrete domains) is an interesting and important problem for future work, it poses a number of significant additional challenges for both deep network model learning as well as Tensorflow-based planning through gradient descent.  We discuss these challenges and possible extensions in more detail in our concluding future work discussion.}, 
$\actionstateconstraints: \mathbb{R}^{|\stateset|} \times 
\mathbb{R}^{|\actionset|} \rightarrow \{\mathit{true},\mathit{false}\}$ is a function that returns 
true if values of action $\bar{A}_{t} = \langle \bar{a}^{1}_{t}, \dots, \bar{a}^{|A|}_{t} \rangle \in \mathbb{R}^{|A|}$ and state $\bar{S}_{t} = \langle \bar{s}^{1}_{t}, \dots, \bar{s}^{|S|}_{t} \rangle \in \mathbb{R}^{|S|}$
variables at time $t$ satisfy global constraints, 
$\transitionfunction: \mathbb{R}^{|\stateset|} \times 
\mathbb{R}^{|\actionset|} \to \mathbb{R}^{|\stateset|}$ denotes the 
transition function between time steps $t$ and $t+1$, 
$\initialstate: \mathbb{R}^{|\stateset|} \rightarrow \{\mathit{true},\mathit{false}\}$ is the initial state constraint indicating which assignment to state variables $\stateset$ is the initial state, and $\goalstate: \mathbb{R}^{|\stateset|} \rightarrow \{\mathit{true},\mathit{false}\}$ represents goal state constraints. 
Finally, $\rewardfunction: \mathbb{R}^{|\stateset|} \times 
\mathbb{R}^{|\actionset|} \rightarrow \mathbb{R}$ denotes 
the reward function\footnote{$Q$ should not be confused with the Q-function in MDPs.  We use $Q$ here since the more standard notation $R$ for immediate reward is used elsewhere for other purposes in this article.}. Given a planning horizon $\horizon$, an optimal 
solution (i.e., an optimal plan) to $\Pi$ is a value assignment to the action variables with values 
$\bar{A}^{t}$ for all time steps $t\in \{1,\dots,H\}$ 
(and state variables with values $\bar{S}^{t}$ 
for all time steps $t\in \{1,\dots,H+1\}$) such that the state variables are updated by the transition function $\transitionfunction(\langle \bar{s}^{1}_{t}, \dots, \bar{s}^{|S|}_{t}, \bar{a}^{1}_{t}, \dots, \bar{a}^{|A|}_{t} \rangle) = \bar{S}_{t+1}$ for all time steps $t\in \{1,\dots, H\}$, initial and goal state constraints are satisfied such that $\initialstate(\bar{S}_{1}) = \mathit{true}$ and $\goalstate (\bar{S}_{H+1}) = \mathit{true}$, and the the total reward function over horizon 
$\horizon$, i.e., $\sum_{t=1}^{\horizon}\rewardfunction(\langle \bar{s}^{1}_{t+1}, \dots, \bar{s}^{|S|}_{t+1}, \bar{a}^{1}_{t}, \dots, \bar{a}^{|A|}_{t} \rangle)$, is maximized.


In many complex problems, it is difficult to model the exact dynamics of 
the complex nonlinear transition function $\transitionfunction$ that governs 
the evolution of states $\stateset$ over the horizon $\horizon$. Therefore in 	
this paper, we do not assume a-priori knowledge of $\transitionfunction$, but 
rather we learn it from data. We limit our model knowledge to a human-specified reward 
function $\rewardfunction$, horizon $\horizon$ and global constraint 
function $\actionstateconstraints$ that specifies whether actions $\actionset$ are 
applicable in state $\stateset$ at time $t$, or not, e.g., the outflow from a reservoir 
must not exceed the present water level, and goal state constraints $\goalstate$.  Given these known components $\stateset, \actionset, 
\actionstateconstraints, \initialstate, \goalstate, \rewardfunction$ of our deterministic factored planning problem, we next discuss how the final component $\transitionfunction$ can be learned as a deep neural network given data.

\section{Neural Network Transition Learning}
\label{transition_learning}

A neural network is a layered, acyclic, directed computational network structure commonly used for supervised learning tasks~\shortcite{goodfellow2016deep}.  
A modern (deep) neural network typically has one (or more) hidden layers, with each hidden layer typically consisting of a linear transformation of its input followed by a nonlinear activation function to produce its output.  Each successive hidden layer provides the ability to learn complex feature structure building on the previous layer(s) that allow it to model nearly arbitrary nonlinear functions.  While most traditional nonlinear activation functions are bounded (e.g., sigmoid or tangent function), simple piecewise linear activation functions such as the rectified linear unit (ReLU)~\cite{relu} have become popular due to their computational efficiency and robustness to the so-called vanishing gradient problem. 

\subsection{Network Structure}

\label{sec:network_structure}

\begin{figure}[t!]
\centering
\includegraphics[width=0.9\linewidth]{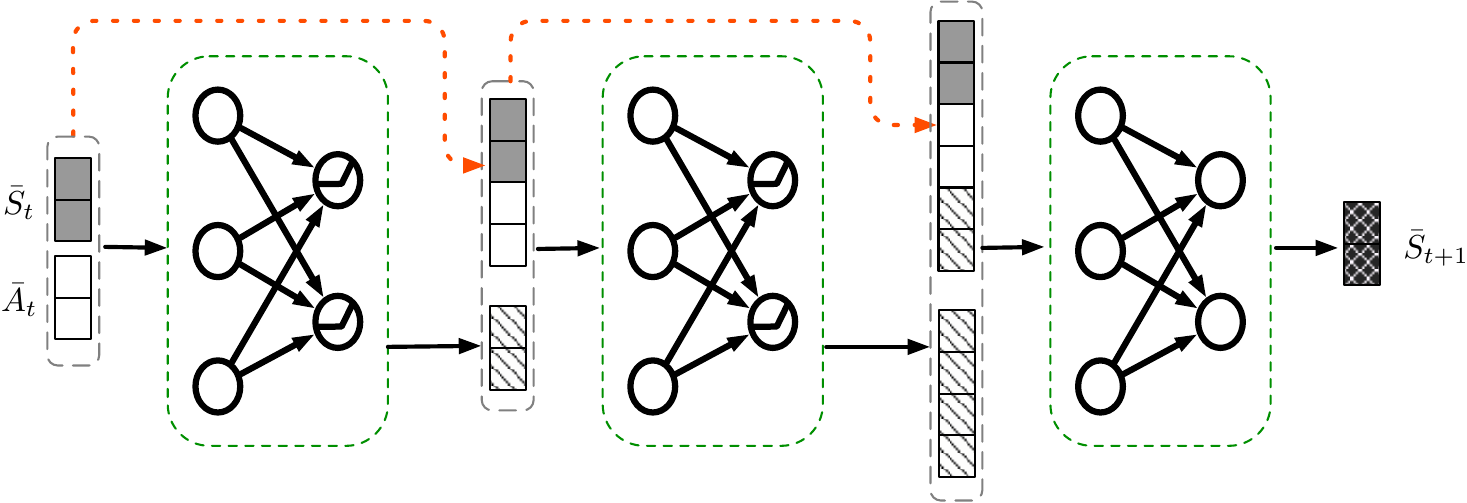}

\caption{An example two-layer neural network transition model.  At left are the input values of state $\bar{S}_{t}$ and action $\bar{A}_{t}$ variables at time $t$, and at far right are the output value state variables $\bar{S}_{t+1}$ at time $t+1$.  The layers of the neural network between this input and output represent the transition function $\transitionfunction(\langle \bar{s}^{1}_{t}, \dots, \bar{s}^{|S|}_{t}, \bar{a}^{1}_{t}, \dots, \bar{a}^{|A|}_{t} \rangle) = \bar{S}_{t+1}$.  In this neural network, black solid arrows indicate  direct connections into each neural network layer, while red dashed lines indicate skip connections that bypass neural network layers and copy the input of the arrow to the output.  Observe that the red dashed skip connections accumulate the concatenation of all previous layers.  Green dashed rounded rectangles represent fully connected layers with an output nonlinearity -- in this paper we use a ReLU nonlinearity.}
 \label{fig:depth}
\end{figure}

In this article, we model the transition function through a deep neural network, where the network takes the current state and current action as input and produces the prediction of the next state as output.  In particular, we use the modified \emph{densely-connected} deep network~\shortcite{huang2017densely}~\footnote{The densely connected network~\cite{huang2017densely} was first proposed for Convolutional Neural Networks but has clear analogues with the fully connected networks we use in this article.} as shown in 
Figure~\ref{fig:depth}, which in comparison to a standard fully connected network, allows direct connections of each layer to the output via skip connections. 
This can be advantageous when a transition function has differing levels 
of nonlinearity (i.e., requiring differing network depths to model), allowing linear components of transitions to pass directly from the input 
to the output layer, while nonlinear components of transitions may pass through one or more of the hidden layers.  We found that it is critical to allow transition function learning to bypass hidden layers when appropriate since, as we will show in Section~\ref{sec:planning_performance}, (a) deeper networks do not necessarily reduce error in transition learning and (b) deeper networks are significantly more time-consuming to optimize with MILP-based solvers.  Thus, having shorter paths enabled by skip connections in dense deep networks can facilitate better learning and faster optimization.


Given a deep neural network configuration with $K$ hidden layers, a hyperparameter $\lambda$, and a dataset $D$ 
consisting of vectors of current state $\bar{\myvec{S}}_n = (\bar{s}^{1}_{n}, \dots, \bar{s}^{|S|}_{n})$, next state $\bar{\myvec{S}}'_n = (\bar{s'}^{1}_{n}, \dots, \bar{s'}^{|S|}_{n})$ 
and action $\bar{\myvec{A}}_n = (\bar{a}^{1}_{n}, \dots, \bar{a}^{|A|}_{n})$ variable values denoted as $D$=$\{ \bar{\myvec{S}}_1{}^\frown \bar{\myvec{A}}_1{}^\frown \bar{\myvec{S}}'_1 , \dots , \bar{\myvec{S}}_N{}^\frown \bar{\myvec{A}}_N{}^\frown \bar{\myvec{S}}'_N  \}$  where ${}^\frown$ denotes the concatenation operation between two vectors, the optimal weights 
$\boldsymbol{W}_k$ for all layers $k\in \{1,\dots,K\}$ can be found by 
minimizing the following prediction error objective:
\begin{align}
&\underset{\mathbf{W}_{k},\myvec{b}_k, k\in \{1,\dots,K\}}
{\text{minimize }} \sum_{n\in \{1,\dots,N\}} \vnorm{\myvec{\gamma} \odot \left[ \bar{\myvec{S}}'_n
-\myvec{S}'_n \right]}_2^2+\lambda\sum_{k\in \{1,\dots,K\}}
\vnorm{\boldsymbol{W}_k}_2^2 \label{ref:learning_obj}\\
&\text{subject to} \nonumber\\
&\myvec{Z}_l=g(\left[ \bar{\myvec{S}}_n{}^\frown\bar{\myvec{A}}_n
{}^\frown\myvec{Z}_1{}^\frown\dots{}^\frown\myvec{Z}_{k-1}\right] \boldsymbol{W}_k^T+
\myvec{b}_k) \quad \forall{k\in \{1,\dots,K-1\}, n\in \{1,\dots,N\}}\label{ref:learning_1}\\
&\myvec{S}'_n = \left[ \bar{\myvec{S}}_n{}^\frown\bar{\myvec{A}}_n{}^\frown
\myvec{Z}_1{}^\frown\dots{}^\frown\myvec{Z}_{K-1}\right]\boldsymbol{W}_{K}^T+
\myvec{b_{K}} \quad \forall{n\in \{1,\dots,N\}}\label{ref:learning_2}
\end{align}
Here, weight matrix $\boldsymbol{W}_k$ and vector of biases $\myvec{b_k}$ for 
each neural network hidden layer $k \in \{1,\ldots,K\}$ are the key parameters being optimized. We use the vector $\myvec{Z}_l$ to represent the intermediate output from the hidden layer $l$.

The objective~\eqref{ref:learning_obj} minimizes squared reconstruction error of transition predictions for the next state in the data, where $\myvec{\gamma}$ denotes a vector of dimensional rescaling weights (defined in the \emph{loss normalization} section below) elementwise multiplied $\odot$ with the vector of residual prediction errors $\left[ \bar{\myvec{S}}'_n -\myvec{S}'_n\right]$.  The objective also contains an $L_2$ regularization term $\vnorm{\boldsymbol{W}_k}_2^2$ with hyperparameter $\lambda \in \mathbb{R}$ to prevent overfitting\footnote{The $L_2$ regularization term on weights $\boldsymbol{W}_k$ is a standard method in the deep learning literature for limiting model complexity to prevent overfitting by penalizing weights with large values, cf. Section 7.1 of \cite{goodfellow2016deep}.}. Constraints (\ref{ref:learning_1})-(\ref{ref:learning_2}) define the nonlinear activation of $\myvec{Z}_k$ for 
hidden layers \mbox{$k\in \{1,\dots,K-1\}$} and outputs $\myvec{S}'_n$, where 
$g$ denotes a nonlinear activation function (specified below).  To reduce notational clutter, we omit subscript $n$ of the hidden layers $\myvec{Z}_k$ though we remark that each unique state $\bar{\myvec{S}}_n$ and action $\bar{\myvec{A}}_n$ variable values naturally lead to a different activation pattern for the $\myvec{Z}_k$.

We use Rectified Linear Units (ReLUs)~\cite{relu} of the form $\mathrm{ReLU}(x)=\max(x,0)$ as the activation function in this paper. ReLUs offer a threefold benefit for our planning tasks:
\begin{enumerate}
\item In comparison to other activation functions, such as the sigmoid or hyperbolic tangent, the ReLU can be trained efficiently and numerically stably since it's derivative is either 1 or 0.
\item Since ReLUs produce a piecewise-linear approximation of the transition function, they permit direct compilation to a set of linear constraints (containing both continuous and integer variables) in Mixed-Integer Linear Programming (MILP) as we will discuss in Section~\ref{MILP}.  Since the number of ReLUs and their constraint-based encoding significantly affects solution time, we do encounter scaling issues with the MILP approach, which we address with our contribution of an alternate gradient-based optimization approach in TF-Plan as we will discuss in Section~\ref{TF}.
\item ReLU activations are robust to the vanishing gradient problem which is advantageous in the context of planning through backpropagation over long horizons, which we will discuss in Section~\ref{sec:long_horizon}.
\end{enumerate}

\subsection{Input Standardization and Loss Normalization}
\label{sec:normalization}

\noindent \paragraph{\bf Input Standardization:} Standardizing input data to have zero mean and unit variance facilitates gradient-based learning used in backpropagation for deep neural networks~\cite{goodfellow2016deep}.  Hence, we standardize inputs 
before feeding them into the neural network, which we found to significantly improve training quality. Specifically, denoting the $i$th neural network input as $\bar{x}^i$, we denote the normalized input as $\hat{x}^i = \frac{\bar{x}^i - \bar{\mu}^i}{\bar{\sigma}^i}$, where $\bar{\myvec{S}}_n{}^\frown \bar{\myvec{A}}_n = (\bar{x}^1_{n},\dots, \bar{x}^{|A|+|S|}_{n})$, and $\bar{\mu}^i$ and $\bar{\sigma}^i$ are respectively the empirical mean and standard deviation of $\bar{x}^i$ in the data.   

After learning, we can modify the weights and biases of the learned neural network to accept unnormalized inputs instead, thus bypassing the need to normalize states and actions during online planning in subsequent sections.  To do this, we define modified weights $\hat{w}_{ij}$ and biases $\hat{b}_j$, respectively for each entry $w_{ij}$ (connecting input $i$ to hidden layer unit $j$) in matrix $\mathbf{W}_k$ and bias $b_j$ (for each hidden layer unit $j$) in vector $\myvec{b}_k$ as follows:
\[\hat{w}_{ij} = \frac{w_{ij}}{\bar{\sigma}^i} \qquad \text{and} \qquad \hat{b}_j= -\sum^{|A|+|S|}_{i=1}\frac{\bar{\mu}^i w_{ij}}{\bar{\sigma}^i} + b_j \, .\]
The full derivation of this result is provided in Appendix~\ref{app:normalization}.

\noindent \paragraph{\bf Loss Normalization:}  While it is critical to standardize the input to improve the learning process, it is also important to normalize the loss applied to each dimension of the output prediction.  

To make the reasons for this more clear, we first begin by observing that it is not uncommon for a next state vector $\myvec{S}'_n$ to contain state variables of highly varying magnitude.  For example, in the Reservoir control problem that we formally define in Section~\ref{sec:domains}, state variables correspond to water volumes of different reservoirs that may have vastly different capacities.  In this case, a uniformly weighted elementwise loss in the objective~\eqref{ref:learning_obj} would lead to most learning effort expended on the largest capacity reservoir at the expense of (relative) accuracy for the smaller reservoirs since the largest reservoir would dominate the squared error. 

To address this loss imbalance, we employ an elementwise loss-weighting vector $\myvec{\gamma}$ on each dimension of the squared loss in the objective~\eqref{ref:learning_obj}.  
Specifically, letting $\bar{\myvec{S}}_{\textit{max}}$ be defined as the elementwise maximum of $\bar{s'}^{i}_1, \dots, \bar{s'}^{i}_N$ of each output dimension $i \in \{1,\dots,|S|\}$ in the data, 
each component of the loss weighting rescales the corresponding error residual by its maximum value observed in the data:
\[\myvec{\gamma} = \frac{1}{\bar{\myvec{S}}_{\textit{max}}} \, . \]
Without the loss normalizing effect of $\myvec{\gamma}$ in the objective~\eqref{ref:learning_obj}, we found it is extremely difficult to learn an accurate predictive model over \emph{all} next state dimensions.


\subsection{Network Regularization and Complexity Reduction} 

\label{sec:hyperparams}


Learning complex functions on limited training data naturally runs the risk of overfitting, which would lead to poor planning performance where planners explore actions and states that may differ substantially from those seen in the training data.  Hence generalization is important and thus in order to combat  
model overfitting, we deployed \emph{Dropout}~\shortcite{srivastava_dropout}, a regularization technique for structured deep networks that randomly ``drops out'' neurons with probability $p$ in order to prevent neurons from co-adapting to each other and memorizing aspects of the training data --- a key factor in deep network overfitting.  More precisely, we use a slight variant known as \emph{Inverted Dropout}~\cite{inverteddropout} that proves to be more stable in post-training generalization and is the default Dropout implementation in many deep learning toolboxes such as Tensorflow.  

In addition to Inverted Dropout for regularizing our deep network, 
it is critical to find the network structure and size that permits good generalization to future state and action scenarios not seen in the training data.  The reasons for this are twofold.  First, minimizing network structure reduces model representation capacity and is by itself a technique to prevent overfitting.  Second, with an eye towards the ultimate planning motivation of this article, we remark that
a smaller network can dramatically reduce the planning computation time.  This is especially true for HD-MILP-Plan, where higher hidden layer width and additional hidden layers result in more \textit{big-M} constraints that substantially increase the computational cost of computing a plan as we discuss and experimentally show in Sections~\ref{MILP} and \ref{experiments}, respectively.  
Structural tuning typically consists of searching over the number of hidden layers as well as the width of these hidden layers and is detailed in 
Section~\ref{experiments}.



\section{Hybrid Deep MILP Planner (HD-MILP-Plan)}
\label{MILP}

In the previous section, we have defined the underlying planning problem addressed in this paper as well as the methodology and structure of deep neural networks used to learn the transition function component.  We now proceed to propose our first planner, HD-MILP-Plan, which can exploit this problem structure and specifically the ReLU structure of our deep net learned transition models by leveraging a compilation to a Mixed-Integer Linear Program (MILP). 

Hybrid~\footnote{The term \emph{hybrid} refers to mixed (i.e., discrete and continuous) action and state spaces as used in MDP literature~\shortcite{kveton2006solving}.} Deep MILP Planner (HD-MILP-Plan) is a two-stage framework for learning and 
optimizing planning problems with piecewise linear transition functions. The first stage of HD-MILP-Plan 
learns the unknown transition function $\transitionfunction$ with densely-connected 
network as discussed previously in Section~\ref{transition_learning}. The learned transition function 
$\tilde{\transitionfunction}$ is then used to construct the learned planning problem 
$\tilde{\Pi} = \langle \stateset, \actionset, \actionstateconstraints, 
\tilde{\transitionfunction}, \initialstate, \goalstate, \rewardfunction \rangle$. 
Given a planning horizon $\horizon$, HD-MILP-Plan compiles the learned planning 
problem $\tilde{\Pi}$ into a MILP and finds an optimal 
plan to $\tilde{\Pi}$ using an off-the-shelf MILP solver. HD-MILP-Plan operates as an 
online planner where actions are optimized over the remaining planning horizon in 
response to sequential state observations from the environment.

We now describe the base MILP encoding of HD-MILP-Plan. Then, we 
strengthen the linear relaxation of our base MILP encoding for solver efficiency.

\subsection{Base MILP Encoding}
\label{ref:base_milp_encoding}

We begin with all notation necessary for the HD-MILP-Plan specification:

\subsubsection{Parameters}

\begin{itemize}[noitemsep,nolistsep]
\item $V_s$ is the value of the initial state variable $s\in S$.
\item $R$ is the set of ReLUs in the neural network.
\item $B$ is the set of learned bias units in the neural network.
\item $O$ is the set of output units in the neural network.
\item $\tilde{w}_{ij}$ denotes the learned weight in the neural network between units $i$ and $j$.
\item $A_f$ is the set of action variables connected to unit $f\in R \cup O$.
\item $S_f$ is the set of state variables connected to unit $f\in R \cup O$.
\item $U_f$ is the set of units connected to unit $f\in R \cup O$.
\item $O_s$ specifies the output unit with linear function that predicts the value of state variable $s\in S$.
\item $M$ is a large constant used in the big-M constraints.
\end{itemize}

\subsubsection{Decision variables}

\begin{itemize}[noitemsep,nolistsep]
\item ${X}^{a}_{t}$ denotes the value of action variable $a\in\actionset$ at time $t$.
\item ${Y}^{s}_{t}$ denotes the value of state variable $s\in\stateset$ at time $t$. 
\item ${P}^{f}_{t}$ denotes the output of ReLU $f\in R$  at time $t$.
\item ${P'}^{f}_{t} = 1$ if ReLU $f\in R$ is activated 
at time $t$, 0 otherwise (i.e., ${P'}^{f}_{t}$ is a Boolean variable).
\end{itemize}

\subsubsection{The MILP Compilation}

Next we define the MILP formulation of our planning optimization problem 
that encodes the learned transition model.
\begin{align}
&\text{maximize }\sum_{t=1}^{\horizon }
\rewardfunction(\langle Y^{s_1}_{t+1},\dots, Y^{s_{|S|}}_{t+1}  ,X^{a_1}_{t},\dots, X^{a_{|A|}}_{t} \rangle)\label{ref:HD_0}\\
&\text{subject to} \nonumber\\
&{Y}^{s}_{1} = V_s \quad \forall{s \in S}\label{ref:HD_1}\\
&\actionstateconstraints(\langle Y^{s_1}_{t},\dots, Y^{s_{|S|}}_{t}, X^{a_1}_{t},\dots, X^{a_{|A|}}_{t}\rangle) \label{ref:HD_2}\\
&\goalstate(\langle Y^{s_1}_{H+1},\dots, Y^{s_{|S|}}_{H+1}\rangle) \label{ref:HD_3}\\
&{P}^{f}_{t} = 1 \quad \forall{f\in B}\label{ref:HD_4}\\ 
&{P}^{f}_{t} \leq M {P'}^{f}_{t} \quad \forall{f \in R}\label{ref:HD_5}\\
&{P}^{g}_{t} \leq M (1-{P'}^{g}_{t}) + \hat{P}^{g}_{t} \quad \forall{g \in R}\label{ref:HD_6}\\
&{P}^{g}_{t}\geq \hat{P}^{g}_{t} \quad \forall{g \in R}\label{ref:HD_7}\\
&{Y}^{s}_{t+1} = \hat{P}^{g}_{t} \quad \forall{g \in O_s, s \in \stateset}\label{ref:HD_8}\\
&\text{where expression }\hat{P}^{g}_{t}=\sum_{f\in U_g}{\tilde{w}_{fg}}{P}^{f}_{t} + \sum_{s\in S_g}{\tilde{w}_{sg}}{Y}^{s}_{t} +\sum_{a\in A_g}{\tilde{w}_{ag}}{X}^{a}_{t} \quad \forall{g \in R}\nonumber\\
&\text{for all time steps }t\in \{1,\dots,\horizon\}\nonumber\text{ except constraints (\ref{ref:HD_1})-(\ref{ref:HD_3}).}\nonumber
\end{align}
In the above MILP, the objective function (\ref{ref:HD_0}) maximizes the sum of 
rewards over a given horizon $\horizon$. Constraint (\ref{ref:HD_1}) connects 
input units of the neural network to the initial state of the planning problem at 
time $t=1$. Constraint (\ref{ref:HD_2}) ensures that global constraints are 
satisfied at every time $t$. Constraint (\ref{ref:HD_3}) ensures output units of 
the neural network satisfy goal state constraints of the planning problem at time 
$t=H+1$. Constraint (\ref{ref:HD_4}) sets all neurons that represent biases equal 
to 1. Constraint (\ref{ref:HD_5}) ensures that a ReLU $f\in R$ is activated if 
the total weighted input flow into $f$ is positive. Constraints 
(\ref{ref:HD_6})-(\ref{ref:HD_7}) together ensure that if a ReLU $f\in R$ is 
active, the outflow from $f$ is equal to the total weighted input flow. 
Constraint (\ref{ref:HD_8}) predicts the values of state variables at time 
$t+1$ given the values of state and action variables, and ReLUs at time $t$ 
using linear activation functions. 

\subsection{Strengthened MILP Encoding}

In the previous encoding, constraints (\ref{ref:HD_5})-(\ref{ref:HD_7}) sufficiently encode 
the piecewise linear activation function of the ReLUs.  However, the positive 
unbounded nature of the ReLUs leads to a poor linear relaxation of the big-M 
constraints, that is, when all boolean variables are 
relaxed to continuous $[0,1]$ in constraints (\ref{ref:HD_5})-(\ref{ref:HD_6}); 
this can significantly hinder 
the overall performance of standard branch and bound MILP solvers that rely 
on the linear relaxation of the MILP for guidance. Consequently, in this section, we 
strengthen our base MILP encoding by preprocessing bounds on state and action 
variables, and with the addition of auxiliary decision variables and linear constraints, to improve its LP relaxation.

In our base MILP encoding, constraints (\ref{ref:HD_5})-(\ref{ref:HD_7}) encode the piecewise 
linear activation function, $\mathrm{relu}(x)=\max(x,0)$, using the big-M 
constraints for each ReLU $f\in R$. We strengthen the linear relaxation of 
constraints (\ref{ref:HD_5})-(\ref{ref:HD_6}) by first finding tighter bounds on the input units of 
the neural network, then separating the input $x$ into its positive 
$x^+$ and negative $x^-$ components. Using these auxiliary variables, 
we augment our base MILP encoding with an additional linear 
inequality in the form of $x^+ \geq \mathrm{relu}(x)$. This inequality 
is valid since the constraints $x = x^+ + x^-$ and 
$\mathrm{relu}(x) = \max(x,0) \leq \max(x^+,0) = x^+$ hold for all
$x^+ \geq 0$ and $x^- \leq 0$. 

\subsubsection{Preprocessing Bounds}
The optimization problems solved to find 
the tightest bounds on the input units of the neural network are as follows.
The tightest lower bounds on action variables can be obtained by solving the 
following optimization problem:
\begin{align}
&\text{minimize } X^{a}_{t} \label{pre_1}\\
&\text{subject to} \nonumber\\
&\text{Constraints (\ref{ref:HD_1})-(\ref{ref:HD_8})} \nonumber
\end{align}
Similarly, the tightest lower bounds on state variables, upper bounds on action 
and state variables can be obtained by simply replacing the expression in the 
objective function (\ref{pre_1}) with $Y^{s}_{t}$, $-X^{a}_{t}$, and $-Y^{s}_{t}$, 
respectively. Given the preprocessing optimization problems have the same 
theoretical complexity as the original learned planning optimization problem (i.e., NP-hard), we limit 
the computational budget allocated to each preprocessing optimization problem 
to a fixed amount, and set the lower and upper bounds on the domains of action 
$a\in \actionset$ and state $s\in \stateset$ decision variables 
${X}^{a}_{t}$, ${Y}^{s}_{t}$ to the best dual bounds found in each respective problem.

\subsubsection{Additional Decision Variables}
The additional decision variables
required to implement our strengthened MILP are as follows:
\begin{itemize}[noitemsep]
\item ${X}^{+,a}_{t}$ and ${X}^{-,a}_{t}$ denote the positive and negative 
values of action variable $a\in\actionset$ at time $t$, respectively.
\item ${Y}^{+,s}_{t}$ and ${Y}^{-,s}_{t}$ denote the positive and negative values of state variable $s\in\stateset$ at time $t$, respectively.
\item ${X'}^{a}_{t} = 1$ if ${X}^{a}_{t}$ is positive at time $t$, 0 otherwise.
\item ${Y'}^{s}_{t} = 1$ if ${Y}^{s}_{t}$ is positive at time $t$, 0 otherwise.
\end{itemize}

\subsubsection{Additional Constraints}
The additional constraints in the strengthened MILP are as follows:
\begin{align}
&{X}^{a}_{t} = {X}^{+,a}_{t} + {X}^{-,a}_{t}\label{strength_1}\\
&{X}^{a}_{t} \leq U^a {X'}^{a}_{t}\label{strength_2}\\
&{X}^{a}_{t} \geq L^a (1-{X'}^{a}_{t})\label{strength_3}\\
&{X}^{+,a}_{t} \leq U^a {X'}^{a}_{t}\label{strength_4}\\
&{X}^{-,a}_{t} \geq L^a (1-{X'}^{a}_{t})\label{strength_5}\\
&\text{for all action variables }a\in \actionset \text{ where }L^a <0 
\text{ and }U^a > 0\text{, time steps }t\in \{1,\dots,\horizon\}\\
&{Y}^{s}_{t} = {Y}^{+,s}_{t} + {Y}^{-,s}_{t}\label{strength_6}\\
&{Y}^{s}_{t} \leq U^s {Y'}^{s}_{t}\label{strength_7}\\
&{Y}^{s}_{t} \geq L^s (1-{Y'}^{s}_{t})\label{strength_8}\\
&{Y}^{+,s}_{t} \leq U^s {Y'}^{s}_{t}\label{strength_9}\\
&{Y}^{-,s}_{t} \geq L^s (1-{Y'}^{s}_{t})\label{strength_10}\\
&\text{for all state variables }s\in \stateset \text{ where }L^s <0 
\text{ and }U^s > 0\text{, time steps }t \in \{1,\dots,\horizon+1\}\nonumber
\end{align}
Here, the ranges $[ L^a, U^a ]$ and $[ L^s, U^s ]$ 
are the lower and upper bounds on the domains of action $a\in \actionset$ 
and state $s\in \stateset$ decision variables ${X}^{a}_{t}$, ${Y}^{s}_{t}$, 
respectively found by solving the preprocessing optimization 
problems. Given constraints (\ref{strength_1})-(\ref{strength_10}), 
constraint (\ref{strength_11}) implements our strengthening constraint which 
provides a valid upper bound on each ReLU $f\in R$.
\begin{align}
&\sum_{s\in S_g, \tilde{w}_{sg} > 0, L^s \geq 0}{\tilde{w}_{sg}}{Y}^{s}_{t} 
+ \sum_{s\in S_g, \tilde{w}_{sg} < 0, U^s \leq 0}{\tilde{w}_{sg}}{Y}^{s}_{t}
+ \sum_{s\in S_g, \tilde{w}_{sg} > 0, L^s <0}{\tilde{w}_{sg}}{Y}^{+,s}_{t}\nonumber\\
&+ \sum_{s\in S_g, \tilde{w}_{sg} < 0, L^s <0}{\tilde{w}_{sg}}{Y}^{-,s}_{t}
+ \sum_{a\in A_g, \tilde{w}_{ag} > 0, L^a \geq 0}{\tilde{w}_{ag}}{X}^{a}_{t} 
+ \sum_{a\in A_g, \tilde{w}_{ag} < 0, U^a \leq 0}{\tilde{w}_{ag}}{X}^{a}_{t}\nonumber\\
&+ \sum_{a\in A_g, \tilde{w}_{ag} > 0, L^a <0}{\tilde{w}_{ag}}{X}^{+,a}_{t} 
+ \sum_{a\in A_g, \tilde{w}_{ag} < 0, L^a <0}{\tilde{w}_{ag}}{X}^{-,a}_{t} 
+ \sum_{f\in U_g\cap R, \tilde{w}_{fg} > 0}{\tilde{w}_{fg}}{P}^{f}_{t} \nonumber\\
&+ \sum_{f\in U_g\cap B}{\tilde{w}_{fg}}{P'}^{g}_{t}\geq {P}^{g}_{t} \label{strength_11}\\
&\text{for all ReLU }g\in R\text{, time steps }t\in \{1,\dots,\horizon\}\nonumber
\end{align}

With this, we conclude our definition of HD-MILP-Plan --- both the base encoding in Section~\ref{ref:base_milp_encoding} and the enhanced encoding in this section for that strengthened the LP relaxation of the big-M constraints for the ReLU activation functions.  We will observe the performance of HD-MILP-Plan and the improvement offered by the strengthened encoding in Section~\ref{experiments}, but we first define an alternative to HD-MILP-Plan for continuous action planning problems in the next section that offers improved scalability but sacrifices provable optimality. 

\section{Tensorflow-based Planner (TF-Plan)}
\label{TF}

While HD-MILP-Plan offers the advantage of provably optimal plans for ReLU deep network learned transition models and discrete and continuous action spaces, we will see in Section~\ref{experiments} that it does face computational constraints that limit its ability to solve large hybrid planning problems.  Hence as an alternative to HD-MILP-Plan, in this section we present a planner based on end-to-end auto-differentiation that we call Tensorflow Planner (TF-Plan). 
TF-Plan represents the planning task as a symbolic recurrent neural network (RNN) architecture with action parameter inputs directly amenable to optimization with GPU-based symbolic toolkits such as Tensorflow and Pytorch. 

The TF-Plan we describe here is an extension of our previous Tensorflow-based tool for planning~\cite{Wu2017} with the following key differences: (1) Instead of compiling the hard-coded transition model of the planning domain into a Tensorflow graph, we learn the transition function through a Tensorflow-based neural network as previously described in Section~\ref{transition_learning} and then freeze the weights of this neural network to obtain a Tensorflow-based model of state transitions. (2) We describe a methodology for handling action constraints used in this work. (3) We evaluate TF-Plan and compare to HD-MILP-Plan on ReLU-based learned neural network transition models that contrast with previous experimental evaluation of TF-Plan on much simpler, manually encoded (non-learned) domain models.

We outline the TF-Plan approach in detail in the following subsections. 

\begin{figure}[t!]
\centering
	\includegraphics[width=1.0\linewidth]{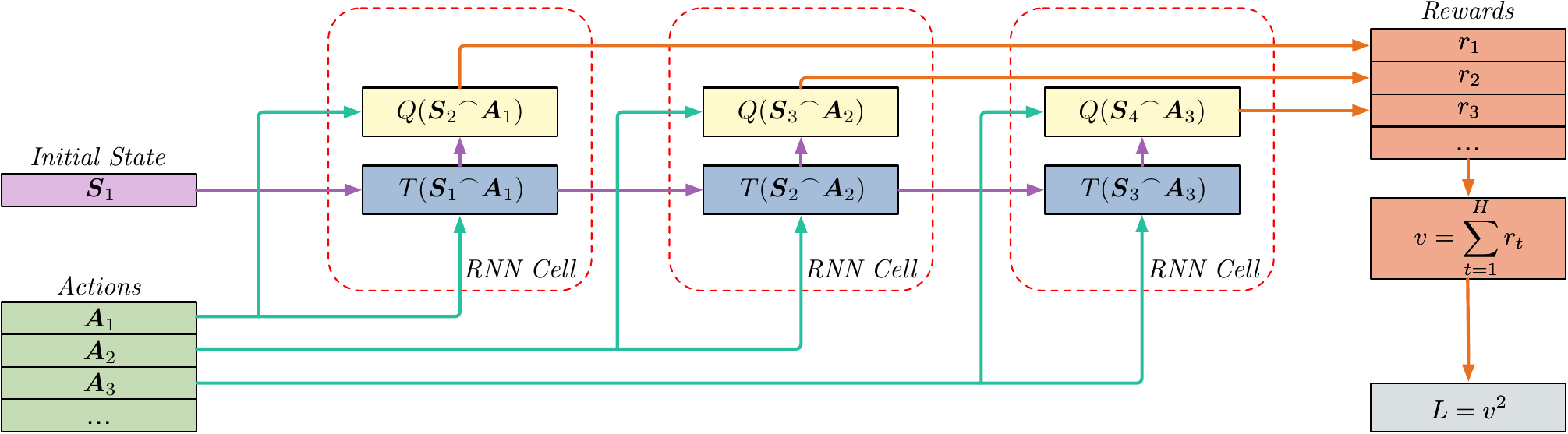}

\caption{A recurrent neural network (RNN) encoding of a planning problem with continuous state and action parameters: A \emph{known} next-state dependent immediate reward ($Q$) and \emph{learned} transition function ($T$) of a discrete time decision-process are embedded in a \emph{custom} RNN cell.  RNN cell inputs correspond to the \emph{fixed} starting state and \emph{free variable} actions to be optimized.  RNN cell outputs correspond to immediate reward and next state. Rewards for each planning instance
are additively accumulated in $v$.  Since the entire specification of objective $L$ is a symbolic representation in Tensorflow with free variable action parameters as inputs, the sequential action plan can be directly optimized via gradient descent using the auto-differentiated representation of $L$.  In practice, we will typically instantiate this RNN with many instances $i$ to support parallel non-convex optimization over multiple initial seeds; here we omit the subscript for the instance $i$ index for notational clarity.  
}
\label{fig:structure}
\end{figure}

\subsection{End-to-end Planning through Backpropagation}

Backpropagation~\shortcite{rumelhart1988learning} is a standard method for optimizing parameters of deep neural networks via gradient descent. With the chain rule of derivatives, backpropagation propagates the derivative of the output error of a neural network back to each of its parameters in a single linear-time pass in the size of the network using what is known as reverse-mode automatic differentiation~\cite{Linnainmaa1970}. Despite its theoretical efficiency, backpropagation in large-scale deep neural networks in practice is still computationally expensive, and it is only with the advent of recent GPU-based symbolic toolkits that efficient backpropagation at a scale and efficiency required to implement end-to-end planning by gradient descent has become possible.

In planning through backpropagation, we reverse the idea of training the parameters of the network to minimize a loss function over sampled data.
Instead, assuming that we have pretrained our deep network transition model, we \emph{freeze} these trained network weights and optimize the inputs (i.e., actions) subject to the fixed transition parameters and an initial state.  This end-to-end symbolic planning framework is demonstrated in the form of a recurrent neural network (RNN) as shown in Figure~\ref{fig:structure}; specifically, given learned transition $\tilde{T}$ and reward $Q$ are piecewise differentiable vector-valued functions, we want to optimize the action input vector $\myvec{A}_t$ for all $t \in \{1,\dots, H\}$ to maximize the accumulated reward value $v = \sum_{t=1}^{H} r_t$ where $r_t = Q(\myvec{S}_{t+1} {}^\frown \myvec{A}_t)$. 
We remark here that we want to optimize \emph{all} actions 
$\myvec{A} = \myvec{A}_1{}^\frown\dots ^\frown\myvec{A}_H$ with respect to a planning loss $L$ (defined shortly as a function of $v$) that we minimize via the following gradient update schema
\begin{equation}
\label{eq:1}
\myvec{A} \leftarrow \myvec{A}-\eta \frac{\partial L}{\partial \myvec{A}},
\end{equation} 
where $\eta$ is the optimization rate and the partial derivatives comprising the gradient based optimization in problem instance $i$,
where we define the total loss $L=\sum_i L_i$ over multiple planning instances $i$. To explain why we have multiple planning instances $i$, we remark that since both transition and reward functions are generally non-convex, optimization on a domain with such dynamics could result in a local minimum or saddle point.  
To mitigate this problem, we randomly initialize actions for a batch of instances and optimize multiple mutually independent planning instances $i$ in parallel (leveraging the GPU implementation of Tensorflow), and eventually return the best-performing action sequence over all instances.

While there are multiple choices of loss function in Auto-differentiation toolkits, we minimize $L_i=v_i^2$ since the cumulative rewards we test in this paper are at most piecewise linear.  We remark that the derivative of a linear function yields a constant value which is not informative in updating actions using the gradient update schema (\ref{eq:1}).  Hence, the optimization of the sum of the squared loss functions $L_i$ has dual effects: (1) it optimizes each problem instance $i$ independently (implemented on the GPU in parallel) and (2) provides fast convergence (i.e., empirically faster than optimizing $v_i$ directly). We remark that simply defining the objective $L$ and the definition of all state variables in terms of predecessor state and action variables via the transition dynamics is enough for auto-differentiation toolkits such as Tensorflow to build the symbolic directed acyclic graph (DAG) of Figure~\ref{fig:structure} representing the objective $L$ and take its gradient with respect to all free action parameters 
using reverse-mode automatic differentiation.

\subsection{Planning over Long Horizons}
\label{sec:long_horizon}

The TF-Plan compilation of a planning problem with a neural network transition model shown in Figure~\ref{fig:structure} reflects the same abstract sequential structure as an RNN that is commonly used in deep learning.  However, we critically remark that we have defined \emph{custom} RNN cells based on the known reward model $Q$ and learned transition model $T$ as opposed to using more well-known RNN cells such as the LSTM~\cite{lstm} that are intended for general sequential learning purposes and do not reflect the specific planning purpose of the TF-Plan architecture.  

Nonetheless, the connection between our RNN representation of hybrid planning in TF-Plan and deep learning with RNNs is not entirely superficial.  A longstanding difficulty with training RNNs lies in the vanishing gradient problem, that is, multiplying long sequences of gradients via the chain rule usually renders the gradients extremely small and negligible for weight updates, especially when using nonlinear activation functions that can be saturated such as a sigmoid.  Because TF-Plan must differentiate through all time steps to optimize action plans, it is similarly subject to the vanishing gradient problem.  

We mitigate the vanishing gradient issue of the RNN in TF-Plan by training the transition function $T$ of each RNN cell in Figure~\ref{fig:structure} with ReLU piecewise linear activations, which has a gradient of either $0$ (ReLU is inactive) or $1$ (ReLU is active) and thus guarantees that the gradient through a path of active ReLUs does not vanish due to the ReLU activations.  We note that reward function $Q$ of each RNN cell does not trigger the vanishing gradient problem since the output $r_t$ of each time step $t$ is directly (additively) connected to the loss function.

\subsection{Handling Action Bound Constraints}

Bounds on actions are common in many planning tasks. For example in the Navigation domain, the distance that the agent can move at each time step is bounded by constant minimum and maximum values.  To handle actions with such range constraints in this planning by backpropagation framework, we use projected stochastic gradient descent.  Projected gradient descent (PGD)~\cite{calamai1987projected} is a method that can handle constrained optimization problems by projecting the parameters (action variables) into a feasible range after each gradient update. Precisely, we clip the values of all action variables $a\in A$ to their feasible range $[L^a,U^a]$ after each epoch of gradient descent:
\[\bar{a} \leftarrow \min(\max(\bar{a}, {L^a}), {U^a})\]
In an online planning setting, TF-Plan only ensures the feasibility of actions with bound constraints using PSGD at time step $t=1$.  TF-Plan does not enforce the remaining global (action) constraints $C$ and goal constraints $G$ during planning.  In general, effectively handling arbitrary constraints in the Tensorflow framework is an open research question beyond the scope of this paper.  Nonetheless, action bounds are among the most common constraints (i.e., the ones we use experimentally) and TF-Plan's initial time step handling of these constraints via PSGD is sufficient to guarantee that only feasible actions are taken during online planning with TF-Plan.

\section{Experimental Results}
\label{experiments}

In this section, we present experimental results that empirically test the performance of both HD-MILP-Plan and TF-Plan on multiple planning domains with learned neural network transition models. 
These experiments focus on continuous action domains since the intent of the paper is to compare the performance of HD-MILP-Plan to TF-Plan on domains where they are both applicable.  To accomplish this task we first present three nonlinear continuous action benchmark domains, namely: Reservoir Control, Heating, Ventilation and Air Conditioning, and Navigation.  Then, we validate the transition learning performance of our proposed ReLU-based densely-connected neural networks with different network configurations in each domain. Finally we evaluate the efficacy of both proposed planning frameworks based on the learned model by comparing them to strong baseline manually coded policies~\footnote{As noted in the Introduction, MCTS and model-free reinforcement learning are not applicable as baselines given our multi-dimensional concurrent \emph{continuous} action spaces.} in an online planning setting. For HD-MILP-Plan, we test the effect of preprocessing to strengthened MILP encoding on run time and solution quality. For TF-Plan, we investigate the impact of the number of epochs on planning quality. Finally, we test the scalability of both planners on large scale domains and show that TF-Plan can scale much more gracefully compared to HD-MILP-Plan.

\subsection{Illustrative Domains} 
\label{sec:domains}

Full RDDL~\cite{Sanner:RDDL} specifications of all domains and instances defined below and used for data generation and plan evaluation in the experimentation are listed in Appendix~\ref{app:domain_descriptions}. \newline
 
\noindent\textbf{Reservoir Control} has a single state variable 
$l^r \in \mathbb{R}$ for each reservoir, which denotes the water level 
of the reservoir $r$ and a corresponding action variable for each $r$ to permit 
a flow $f^r \in [0,r^{\max}]$ from reservoir $r$ (with maximum allowable 
flow $r^{\max}$) to the next downstream reservoir. The transition is 
a nonlinear function due to the evaporation ${e^r}$ from each reservoir 
$r$, which is defined by the formula 
\[e^r_t = (1.0/2.0)\cdot\sin((1.0/2.0)\cdot l^r_t)\cdot0.1 ,\]
 and the water level transition function is 
\[ l_{t+1}^r= l_{t}^r+\sum_{r_{\mathit{up}}} 
f^{r_{\mathit{up}}}-{f}_t^r-e_t^r ,\] 
where $f^{r_\mathit{up}}$ ranges over all upstream reservoirs of $r$ with bounds $0 \leq f^r \leq l^r$. 
The reward function minimizes the total absolute deviation from a 
desired water level, plus a constant penalty for having water 
level outside of a safe range (close to empty or overflowing), which 
is defined for each time step $t$ by the 
expression
\begin{align*}
-\sum_{r} 
(0.1\cdot\Big|((m^r + n^r)/2.0)-l_{t+1}^r\Big| + 100\cdot \max(m^r - l_{t+1}^r, 0 ) + 5\cdot \max(l_{t+1}^r - n^r, 0 )),
\end{align*}
where $m^r$ and $n^r$ define the upper and and lower desired ranges 
for each reservoir $r$. We report the results on small instances with 3 and 4 
reservoirs over planning horizons $\horizon = 10, 20$, and large instances with 10 reservoirs over planning horizons $\horizon = 10, 20$. \newline

\noindent\textbf{Heating, Ventilation and Air Conditioning}~\cite{agarwal2010} 
has a state variable $p^r \in \mathbb{R}$ denoting the temperature 
of each room $r$ and an action variable $b^r \in [0,b^{\max}]$ for sending heated air 
to each room $r$ (with maximum allowable volume $b^{\max}$) via vent actuation. 
The bilinear transition function is then   
\[p_{t+1}^r=p_{t}^r + (\Delta t/{C}^{r}) ({b^r} +
\sum_{r'} ({p}_{t}^{r'}-p_{t}^r)/R^{rr'}) ,\]
where ${C}^{r}$ is the heat capacity of rooms, $r'$ represents 
an adjacency predicate with respect to room $r$ and $R^{rr'}$ represents a 
thermal conductance between rooms.
The reward function minimizes the total absolute deviation from a desired temperature for
all rooms plus a linear penalty for having temperatures outside of a 
range plus a linear penalty for heating air with cost $k$, and 
is defined for each time step $t$ by the expression
\begin{align*} 
-\sum_{r} 
(10.0\cdot|((m^r + n^r)/2.0)-p_t^r| + k b^r + 0.1\cdot(\max(p_t^r - n^r, 0 ) + \max(m^r - p_t^r, 0 )).
\end{align*}
We report the results on small instances with 3 and 6 
rooms over planning horizons $\horizon = 10, 20$, and a large instance with 60 rooms over planning horizon $\horizon = 2$. 
\newline

\noindent \textbf{Navigation} is designed to test
learning of a highly nonlinear transition function and has a vector of state variables for the 
2D location of an agent ${\myvec{P}}_{t+1} = (p_t^x, p_t^y)$ and 
a 2D action variable vector intended nominally to move the agent $\Delta \myvec{P}_{t} = (\Delta p_t^x, \Delta p_t^y)$ (with minimum and maximum movement boundaries $[\Delta p^{\min}, \Delta p^{\max}]$). 
The new location ${\myvec{P}}_{t+1}$ is a nonlinear function of the current 
location $ {\myvec{P}}_t$ (with minimum and maximum maze 
boundaries $[p^{\min},p^{\max}]$) due to higher slippage in the center of
the domain where the transition function is
\[{\myvec{P}}_{t+1} = \myvec{P}_{t} + \Delta 
\myvec{P}_{t} \cdot2.0/(1.0+\exp(-2\cdot\Delta d^p_{t}))-0.99 .\] 
where $\Delta d^p_{t}$ is the Euclidean distance from $\myvec{P}_{t}$ to the 
center of the domain. The reward function minimizes the total Manhattan 
distance from the goal location, which is defined for each time step $t$ 
by the expression
\[ 
-\sum_{d\in\{x,y\}} |g^d - p_t^d|,\]
where $g^d$ defines the goal location for dimension $d$. We report the results on 
small instances with maze sizes 8-by-8 (i.e., $p^{\min} = 0, p^{\max} = 8$) and 10-by-10 (i.e., $p^{\min} = 0, p^{\max} = 10$) over planning horizons $\horizon = 8, 10$, and a large instance with minimum and maximum movement boundaries $[\Delta p^{\min} = -0.5, \Delta p^{\max}= 0.5]$ over planning horizon $\horizon = 20$.

\subsection{Transition Learning Performance}
\label{sec:transition_learning}

In Table~\ref{tab:table1}, we show the held-out test data mean squared error (MSE) of the best configuration of neural network architectures for three instances of the previously defined planning domains.  We train all neural networks using $10^5$ data samples from simulation using a simple stochastic exploration policy. 
As standard for deep network training, we randomly permute the data order for each epoch of gradient descent.
80\% of the sampled data was used for training --- with hyperparameters tuned on a subset of 20\% validation data of the training data --- and 20\% of the sampled data was held out for the test evaluation.   
We applied the  RMSProp~\shortcite{hinton2012neural} optimizer over 200 epochs.
Since densely-connected networks~\cite{huang2017densely}  discussed and illustrated in Section~\ref{sec:network_structure} strictly dominated the performance of non-densely-connected networks, we only report the results of the densely-connected network.
Throughout all experiments, we fixed the dropout parameter $p=0.1$ (cf. Section~\ref{sec:hyperparams}) at \emph{all} hidden layers since deviations from this value rarely improved generalization performance by a significant amount.
We tuned the number of hidden layers in the set of $\{0 \textrm{ (linear)}, 1, 2 \}$ and the number of neurons for each layer in the set of $\{8, 16, 32, 64, 128\}$; for a given layer size, we chose the minimal number of neurons that performed best on validation data within statistical significance.

Overall, we see that Reservoir and HVAC can be accurately learned with one hidden layer since an additional layer hurt generalization performance on the test data.  Navigation benefits from having two layers owing to the complexity of its nonlinear transition and approximation with ReLU activations.  The network with the lowest MSE is used as the deep neural network model for each domain in the subsequent planning experiments.

\begin{figure*}[t!]
    \begin{subfigure}{0.33\textwidth}
  	\includegraphics[width=1\linewidth]{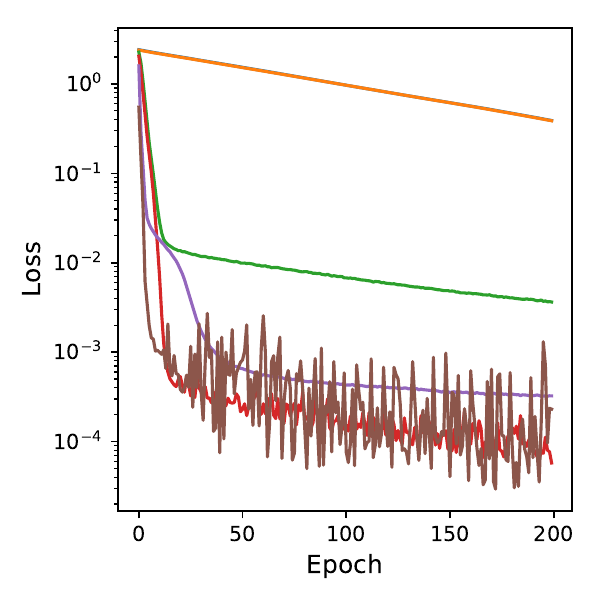}
    \caption{Reservoir with 4 reservoirs}
  	\end{subfigure}
    \begin{subfigure}{0.33\textwidth}
  	\includegraphics[width=1\textwidth]{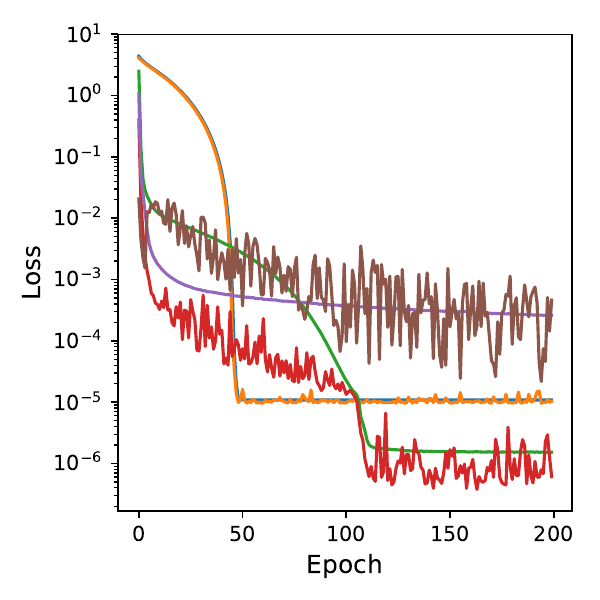}
    \caption{HVAC with 3 rooms}
    \end{subfigure}
  	\begin{subfigure}{0.33\textwidth}
  	\includegraphics[width=1\linewidth]{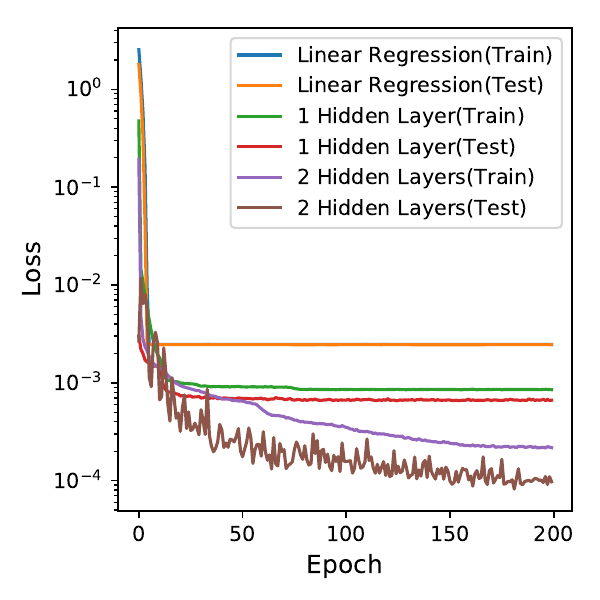}
    \caption{Navigation with 10-by-10 maze}
  	\end{subfigure}
  	\begin{subfigure}{0.33\textwidth}
  	\includegraphics[width=1\linewidth]{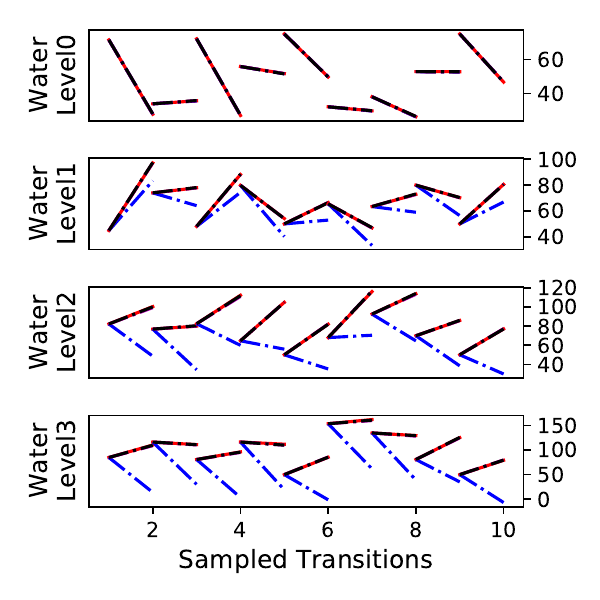}
    \caption{Reservoir with 4 reservoirs}
  	\end{subfigure}
    \begin{subfigure}{0.33\textwidth}
  	\includegraphics[width=1\textwidth]{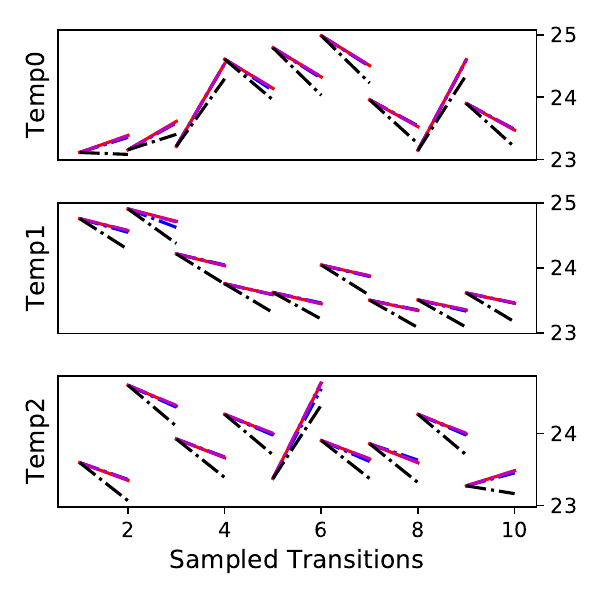}
    \caption{HVAC with 3 rooms}
    \end{subfigure}
  	\begin{subfigure}{0.33\textwidth}
  	\includegraphics[width=1\linewidth]{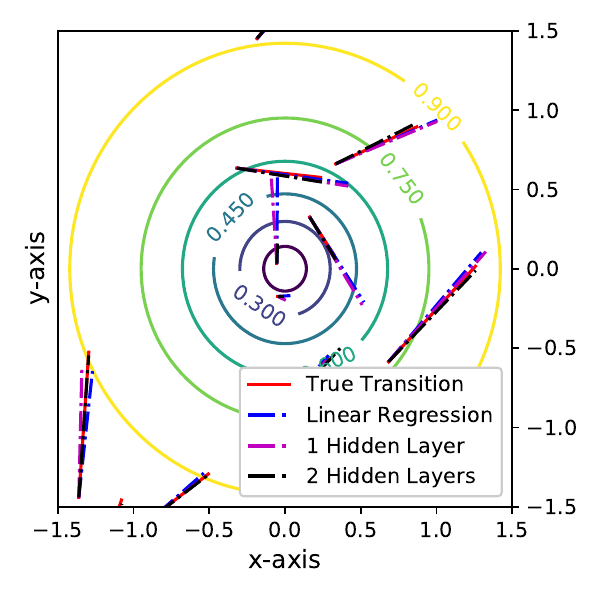}
    \caption{Navigation with 10-by-10 maze}
  	\end{subfigure}
  \caption{Visualization of training performance of different neural network configurations over three domain instances. Figures (a)-(c) represent the loss curves over training epochs for the three domains we introduced in the paper. The y-axis is in logarithmic scale. In figure (a), we observe that while MSE of 1 hidden layer and 2 hidden layer networks look similar, the former has smaller variance. In figures (b)-(c), we observe that 1 hidden layer network and 2 hidden layer network have smallest MSEs, respectively. Figures (d)-(f) represent the learning quality comparisons of linear, 1 hidden layer and 2 hidden layer networks for 3 domains. For different domain variables, we show different starting points and true transitions after a period of time. The dashed lines are the different neural network learned approximations. 
  }
  \label{fig:learning_quality}
\end{figure*}

Figure~\ref{fig:learning_quality} visualizes the training performance of different neural network configurations over three domain instances. Figures~\ref{fig:learning_quality} (a)-(c) visualize the loss curves over training epochs for three domain instances. In Reservoir, we observe that while both 1 and 2 hidden layer networks have similar MSE values, the former has much smaller variance. In HVAC and Navigation instances, we observe that 1 and 2 hidden layer networks have the smallest MSE values, respectively. Figures~\ref{fig:learning_quality} (d)-(f) visualize the performance of learning transition functions with different number of hidden layers. We observe that Reservoir needs at least one hidden layer to overlap with ground truth whereas HVAC is learned well with all networks (all dashed lines overlap) and Navigation only shows complete overlap (especially near the center nonlinearity) for a two layered neural network. All of these results mirror the MSE comparisons in Table~\ref{tab:table1} thus providing both empirical and intuitive evidence justifying the best neural network structure selected for each domain.

\begin{table}[t!]
  \centering
  \caption{Mean Squared Error (MSE) Table for all domains and network configurations with 95\% confidence intervals; three significant digits are shown for MSE.}
  \resizebox{0.95\textwidth}{!}{%
  \begin{tabular}{cccc}
    \toprule
    Domain & Linear & 1 Hidden & 2 Hidden \\
    \midrule
    Reservoir (instance with 4 reservoirs)& 46500000 $\pm$487000 & \textbf{343000 $\pm$7210} & 653000 $\pm$85700 \\ 
    HVAC (instance with 3 rooms)& 710$\pm$2.3 & \textbf{520$\pm$54} & 75200$\pm$7100\\
    Navigation (instance with 10 by 10 maze) & 30400$\pm$9.8 & 9420$\pm$29 & \textbf{1940$\pm$50} \\ 
    \bottomrule
  \end{tabular}
  }
  \label{tab:table1}
\end{table}

\begin{figure*}[t!]
	\begin{subfigure}{0.33\textwidth}
  	\includegraphics[width=1\linewidth]{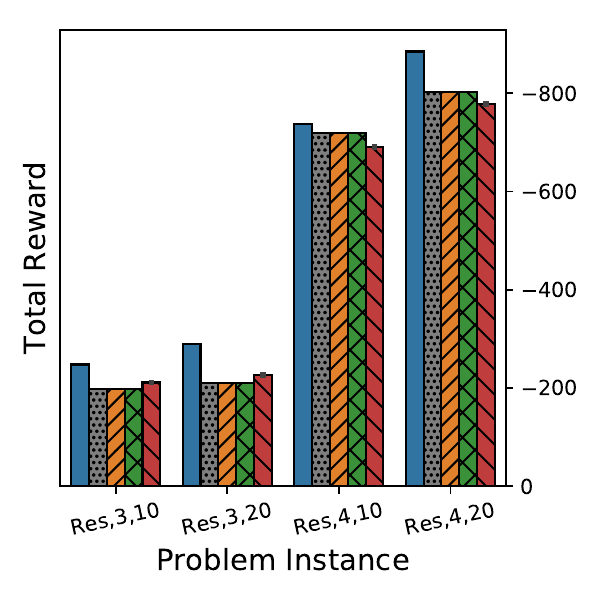}
    \caption{Reservoir}
  	\end{subfigure}
	\begin{subfigure}{0.33\textwidth}
  	\includegraphics[width=1\textwidth]{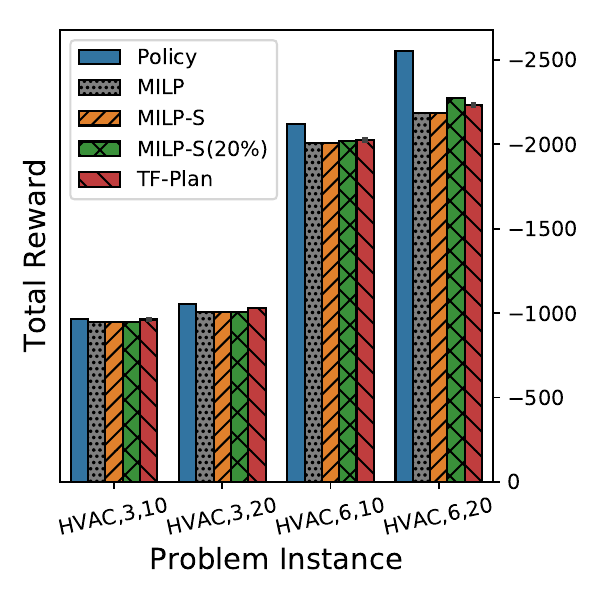}
    \caption{HVAC}
    \end{subfigure}
  	\begin{subfigure}{0.33\textwidth}
  	\includegraphics[width=1\linewidth]{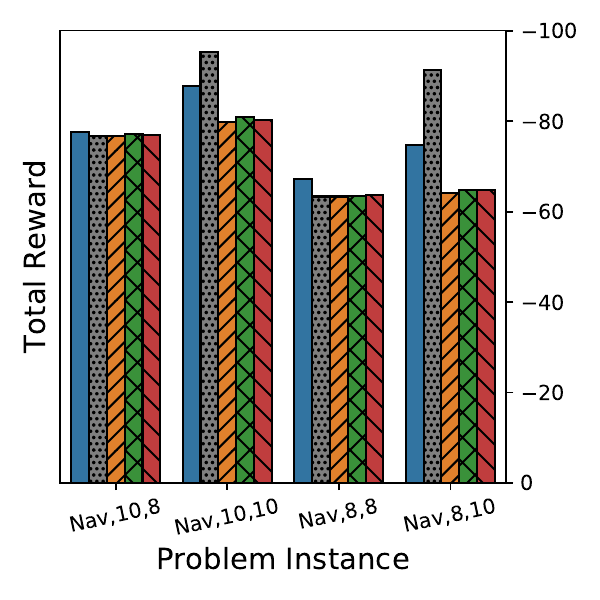}
    \caption{Navigation}
  	\end{subfigure}\\
  	\begin{subfigure}{0.33\textwidth}
  	\includegraphics[width=1\linewidth]{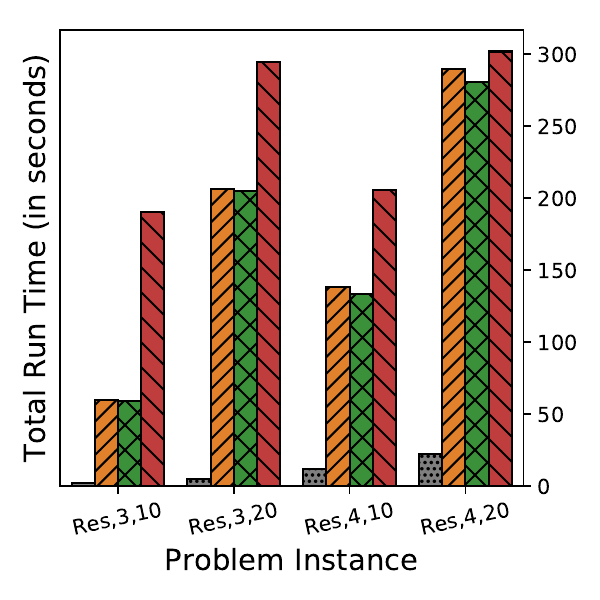}
    \caption{Reservoir}
  	\end{subfigure}
    \begin{subfigure}{0.33\textwidth}
  	\includegraphics[width=1\textwidth]{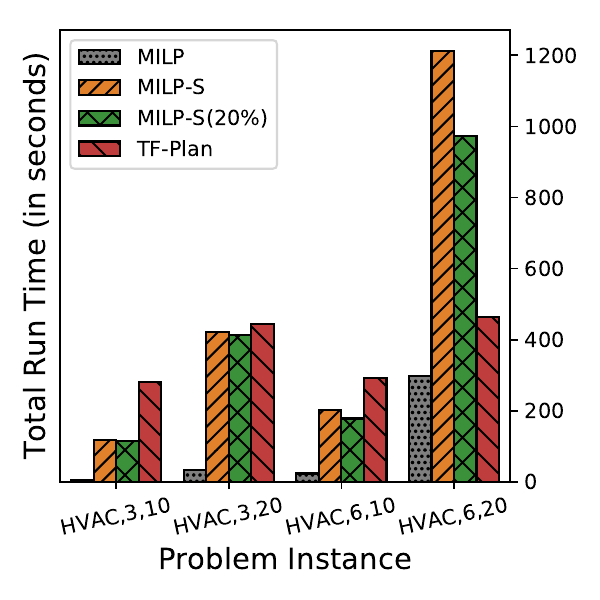}
    \caption{HVAC}
    \end{subfigure}
  	\begin{subfigure}{0.33\textwidth}
  	\includegraphics[width=1\linewidth]{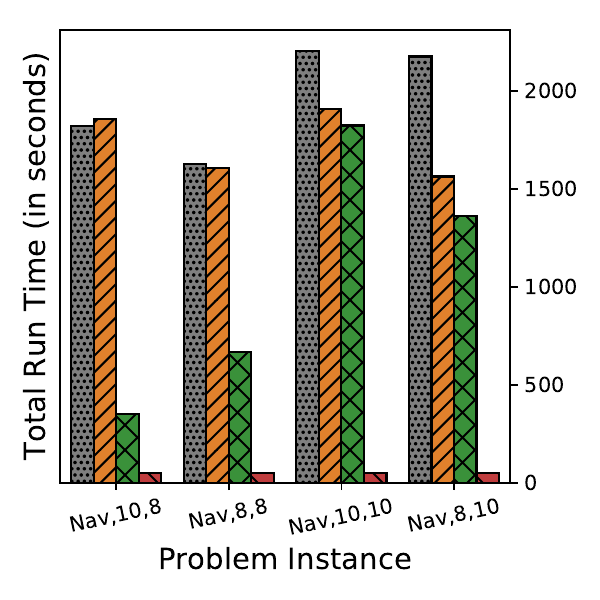}
    \caption{Navigation}
  	\end{subfigure}
  \caption{Overall planning performance comparison.  The domain notation shown in the bar labels of each figure correspond to 
$(\textsc{Domain Name},\textsc{Size},\textsc{Horizon})$.  (a)-(c) visualizes the total \emph{negative} reward comparison (lower is better) between the proposed methods and domain-specific rule-based planning. In addition, we show results for HD-MILP-Plan solved up to 20\% duality gap to provide a reference to the solution quality of TF-Plan. Note that rewards in this paper represent costs to minimize 
and therefore \emph{lower} total negative reward indicates \emph{better} performance. 
The handcoded 
policies were strong baselines intended to be near-optimal, but we see 
greater performance separation as the domains become more nonlinear 
(most notably Reservoir and Navigation) and the optimal policies 
become harder to manually encode. Figures (d)-(f) visualizes the timing comparison among base MILP algorithm (gray bar), 
the strengthened MILP algorithm (orange bar), strengthened MILP with 20\% gap (green bar), and the TF-Plan (red bar). As the domains become more nonlinear 
(i.e., as measured by the learning quality of each domain as presented in 
Table~\ref{tab:table1}) and the deep network depth increases, the strengthened 
MILP encoding begins to dominate the base encoding. Deep network depth 
impacts performance of the base MILP more than problem size.
}
  \label{fig:planning_performance}
\end{figure*}

\begin{figure*}[t!]
    \begin{subfigure}{0.33\textwidth}
  	\includegraphics[width=1\linewidth]{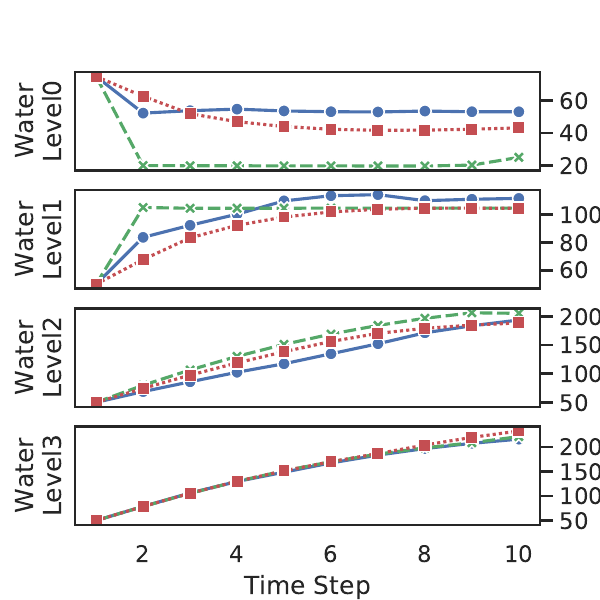}
    \caption{Reservoir with 4 reservoirs}
  	\end{subfigure}
    \begin{subfigure}{0.33\textwidth}
  	\includegraphics[width=1\textwidth]{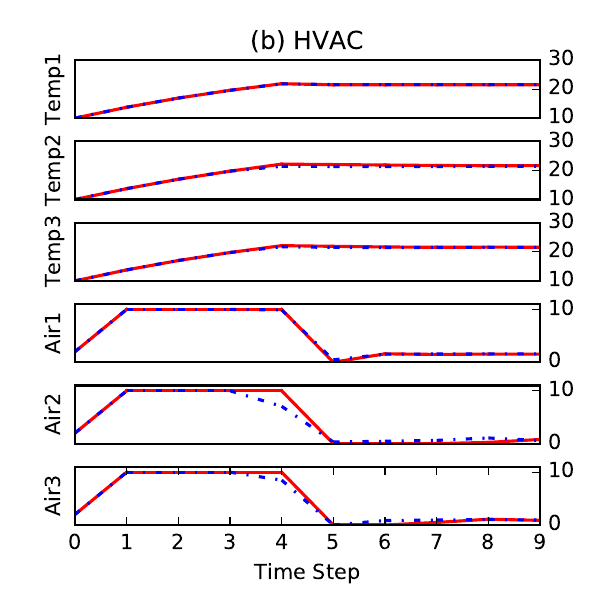}
    \caption{HVAC with 6 rooms}
    \end{subfigure}
  	\begin{subfigure}{0.33\textwidth}
  	\includegraphics[width=1\linewidth]{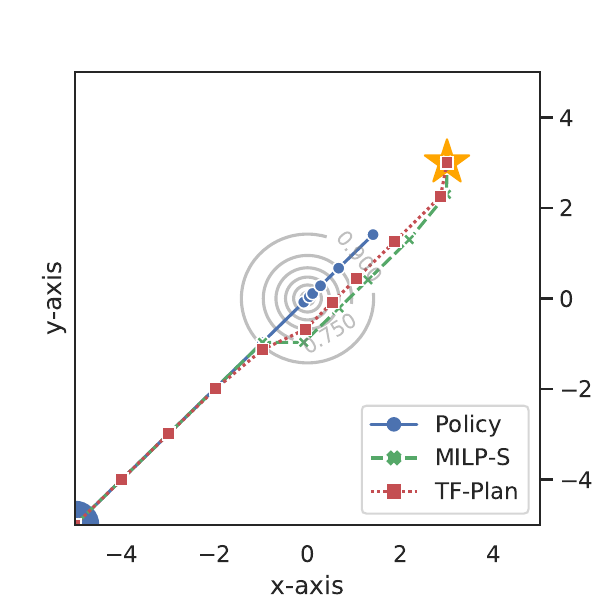}
    \caption{Navigation with 10-by-10 maze}
  	\end{subfigure}
  \caption{Planning behavior comparison between the manually encoded rule-based policy 
(Blue), HD-MILP-Plan (Green) and TF-Plan (Red).  Compared to the strong manually coded policies, both HD-MILP-Plan and TF-Plan
make more subtle nonlinear deviations in comparison to the 
manual policy (Blue) to better optimize the overall
objective as shown in Figures~\ref{fig:planning_performance} (a)-(c).
}
\label{fig:behaviour}
\end{figure*}

\subsection{Planning Performance}
\label{sec:planning_performance}
In this section, we investigate the effectiveness of planning with HD-MILP-Plan and TF-Plan to plan for the original planning problem $\Pi$ through optimizing the learned planning problem $\tilde{\Pi}$ in an online planning setting. We optimized the MILP 
encodings using IBM ILOG CPLEX 12.7.1 with eight threads and a 1-hour total time limit per problem instance on a MacBookPro with 2.8 GHz Intel Core i7 16 GB memory. We optimize TF-Plan through Tensorflow 1.9 with an Nvidia GTX 1080 GPU with CUDA 9.0 on a Linux system with 16 GB memory.\footnote{Due to the fact that CPLEX and Tensorflow leverage different hardware components (GPUs in the case of Tensorflow), the run-times reported are machine-specific and may vary with other machine architecture  configurations and components.} We connected both planners with the  RDDLsim~\cite{Sanner:RDDL} domain simulator 
and interactively solved multiple problem instances with different sizes and horizon lengths (as described in Figure~\ref{fig:planning_performance}). In order to assess the approximate solution quality of TF-Plan, we also report HD-MILP-Plan with 20\% duality gap. The results reported for TF-Plan, unless otherwise stated, are based on fixed number of epochs for each domain where TF-Plan used 1000 epochs for Reservoir and HVAC, and 300 epochs for Navigation.

\subsubsection{Comparison of Planning Quality}

In Figures~\ref{fig:planning_performance} (a)-(c), we compare the planning qualities of the domain-specific policies (blue), the base MILP model (gray), MILP model with preprocessing and strengthening constraints solved optimally (orange), MILP model with preprocessing and strengthening constraints solved up to 20\% duality gap (green) and TF-Plan.

In Figure~\ref{fig:planning_performance} (a), we compare HD-MILP-Plan and TF-Plan to a rule-based local Reservoir planner, which measures the water level in reservoirs, and sets outflows to release water above a pre-specified median level of reservoir capacity. In this domain, we observe an average of 15\% increase in the total reward obtained by the plans generated by HD-MILP-Plan in comparison to that of the rule-based local Reservoir planner. Similarly, we find that TF-Plan outperforms the rule-based local Reservoir planner with a similar percentage though does not always beat HD-MILP-Plan. However, we observe that TF-Plan outperforms both the rule-based policy and HD-MILP-Plan on the larger Reservoir 4 domains. We investigate this outcome further in Figure~\ref{fig:behaviour} (a). We find that the plan returned by HD-MILP-Plan incurs more penalty due to the noise in the learned transition model, where the plan attempts to distribute water to multiple reservoirs and obtain higher reward. As a result, the actions returned by HD-MILP-Plan break the safety threshold and receive additional penalty. Thus, HD-MILP-Plan incurs more cost than TF-Plan.

In Figure~\ref{fig:planning_performance} (b), we compare HD-MILP-Plan and TF-Plan to a rule-based local HVAC policy, which turns on the air conditioner anytime the room temperature is below the median value of a given range of comfortable temperatures [20,25] and turns off otherwise. 
While the reward (i.e., electricity cost) of the proposed models on HVAC 3 rooms are almost identical to that of the locally optimal HVAC policy, we observe significant performance improvement on HVAC 6 rooms settings, which suggests the advantage of the proposed models on complex planning problems, where the manual policy fails to track the temperature interaction among the rooms. Figure~\ref{fig:behaviour} (b) further demonstrates the advantage of our planners where the room temperatures controlled by the proposed models are identical to the locally optimal policy with 15\% less power usage.

Figure~\ref{fig:planning_performance} (c) compares HD-MILP-Plan and TF-Plan to a greedy search policy, which uses a Manhattan distance-to-goal 
function to guide the agent towards the direction of the goal (as visualized by Figure~\ref{fig:behaviour} (c)). The pairwise comparison of the total rewards obtained 
for each problem instance per plan shows that the proposed models can outperform the manual policy up to 15\%, as observed in the problem instance Navigation,10,8 in Figure~\ref{fig:planning_performance} 
(c). The investigation of the actual plans, as visualized by Figure~\ref{fig:behaviour} (c), shows that the local policy ignores the nonlinear region in the middle, and tries to reach the goal directly, which cause the plan to be not able to reach the goal position with given step budget.
In contrast, both HD-MILP-Plan and TF-Plan can find plans that move 
around the nonlinearity and successfully reach the goal state, which shows their ability to model the nonlinearity and find plans that are near-optimal with respect to the learned model over the complete horizon $\horizon$.

Overall we observe that in 10 out of 12 problem instances, the solution quality 
of the plans generated by HD-MILP-Plan and TF-Plan are significantly better than the total reward obtained by the plans generated by the respective domain-specific human-designed policies. Further we find that the quality of the plans generated by TF-Plan are between the plans generated by i) HD-MILP-Plan solved to optimality, and ii) HD-MILP-Plan solved to 20\% duality gap, thus indicating the overall strong approximate performance of TF-Plan.

\subsubsection{Comparison of Run Time Performance}
\label{timing}

In Figures~\ref{fig:planning_performance} (d)-(f), we compare the run time performances of the base MILP model (gray), MILP model with preprocessing and strengthening constraints solved optimally (orange), MILP model with preprocessing and strengthening constraints solved upto 20\% duality gap (green) and TF-Plan.
Figure~\ref{fig:planning_performance} (f) shows significant 
run time improvement for the strengthened encoding over the base MILP 
encoding, while Figures~\ref{fig:planning_performance} (d)-(e) show otherwise. Together with the results presented in Figures~\ref{fig:planning_performance} (a)-(c), we find that domains that utilize neural networks with only 1 hidden layer (e.g., HVAC and Reservoir) do not benefit from the additional fixed computational expense of preprocessing. In contrast, domains that require deeper neural networks (e.g., Navigation) benefit from the additional computational expense of preprocessing and strengthening. Over three domains, we find that TF-Plan significantly outperforms HD-MILP-Plan in all Navigation instances, performs slightly worse in all Reservoir instances, and performs comparable in HVAC instances.

\subsubsection{Effect of Training Epochs for TF-Plan on Planning Quality}

\begin{figure*}[t!]
    \begin{subfigure}{0.33\textwidth}
  	\includegraphics[width=1\linewidth]{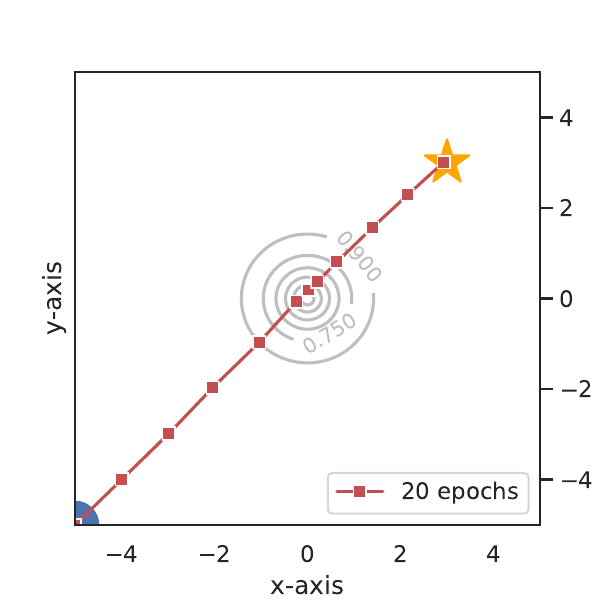}
    \caption{20 epochs}
  	\end{subfigure}
    \begin{subfigure}{0.33\textwidth}
  	\includegraphics[width=1\textwidth]{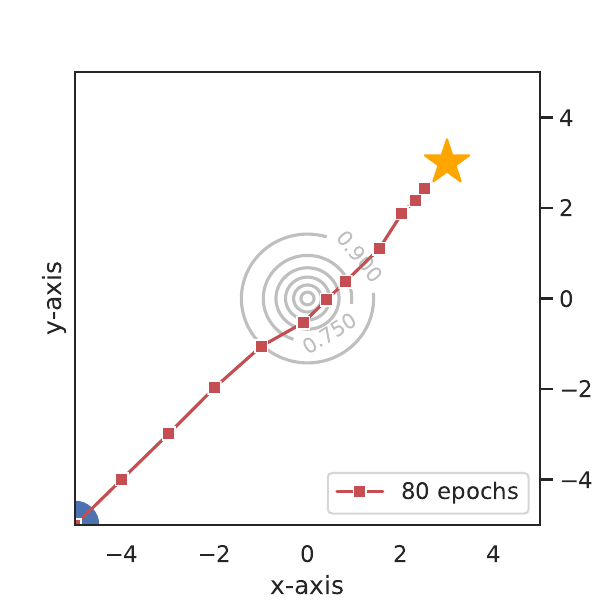}
    \caption{80 epochs}
    \end{subfigure}
  	\begin{subfigure}{0.33\textwidth}
  	\includegraphics[width=1\linewidth]{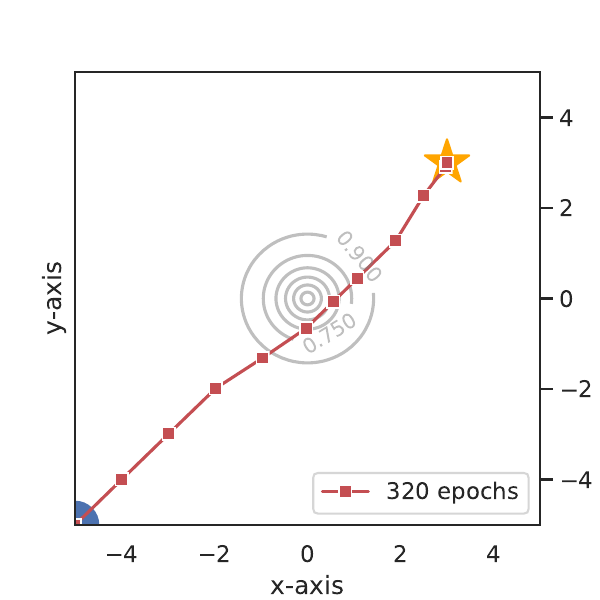}
    \caption{320 epochs}
  	\end{subfigure}
  \caption{Performance comparison using differing numbers of epochs when optimizing with TF-Plan on the Navigation domain. (a) 20 epochs allows the model to find a brute force solution that finds a direct, but highly suboptimal path to the goal that passes through the heart of the central deceleration zone.  (b) 80 epochs allows the model to avoid the deceleration zone, though it take suboptimally large or small steps.  (c) 320 epochs shows a near-optimal result where the agent skirts the deceleration zone and spaces its steps fairly equally.  Furthermore, 320 epochs reaches the goal within 10 steps while 20 epochs requires 11 steps. 
  }
  \label{fig:epochs}
\end{figure*}

To test the effect of the number of optimization epochs on the solution quality, we present results on 10-by-10 Navigation domain for a horizon of 10 with different epochs. Figure~\ref{fig:epochs} visualizes the increase in solution quality as the number of epochs increase where Figure~\ref{fig:epochs} (a) presents a low quality plan found similar to that of the manual-policy with 20 epochs, Figure~\ref{fig:epochs} (b) presents a medium quality plan with 80 epochs, and Figure~\ref{fig:epochs} (c) presents a high quality plan similar to that of HD-MILP-Plan with 320 epochs.

\begin{figure*}[t!]
	\begin{subfigure}{0.33\textwidth}
  	\includegraphics[width=1\linewidth]{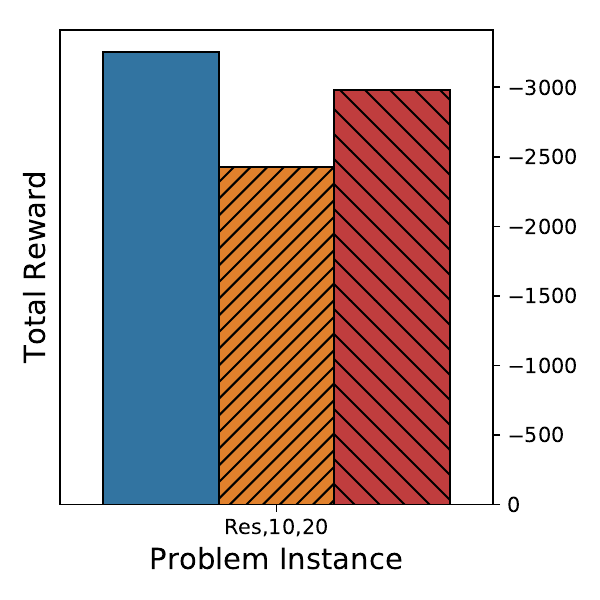}
    \caption{Reservoir}
  	\end{subfigure}
	\begin{subfigure}{0.33\textwidth}
  	\includegraphics[width=1\textwidth]{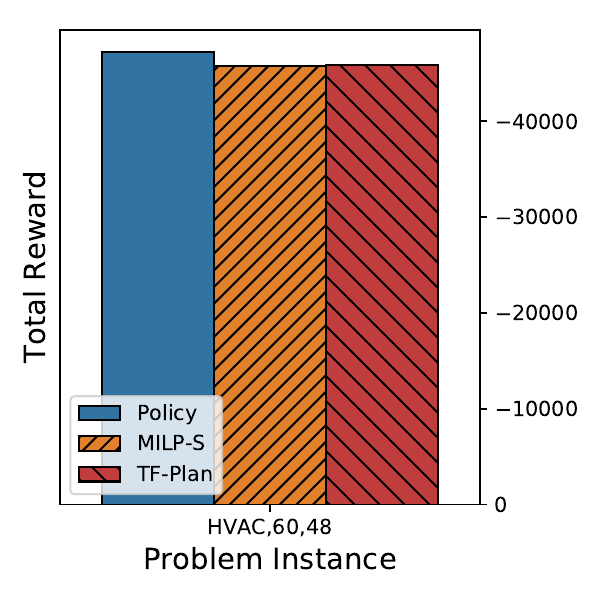}
    \caption{HVAC}
    \end{subfigure}
  	\begin{subfigure}{0.33\textwidth}
  	\includegraphics[width=1\linewidth]{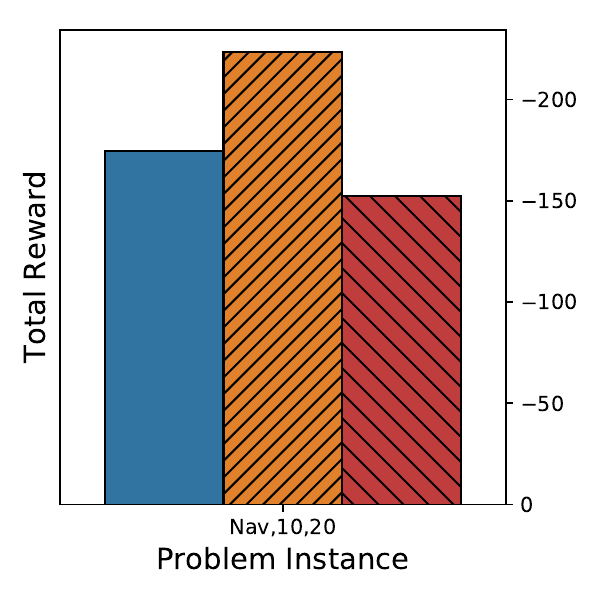}
    \caption{Navigation}
  	\end{subfigure}
  \caption{The total reward comparison between the proposed methods and the (strong baseline) domain-specific rule-based planning on larger problem instances. Since the y-axis is a negative cost, lower (less cost) is better.  In particular, we show TF-Plan (red) scales better by consistently finding plans better than the rule-based policy (blue). Further, we show HD-MILP-Plan (orange) outperforms the other planners in two out of three domains (i.e., Reservoir and HVAC), but performs poorly in the Navigation domain due to its deeper neural network structure and longer planning horizon that requires a large MILP encoding.}
  \label{fig:scale}
\end{figure*}

\begin{figure}[t!]
\centering
  \includegraphics[width=0.66\linewidth]{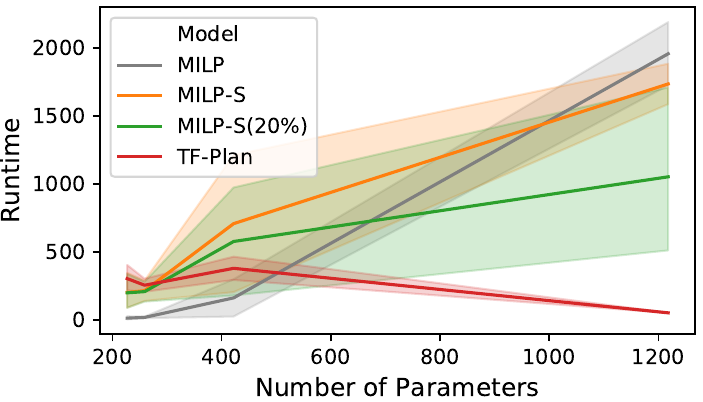}
  \caption{Visualization of run times as a function of problem sizes (specifically the number of variables in the encoding), where problem size is a function of horizon $\horizon$, number of parameters in the learned model, and number of neural network layers. 95\% confidence intervals on running times are shown.
  }
  \label{fig:time_size}
\end{figure}

\subsubsection{Scalability Analysis on Large Problem Instances}
To test the scalability of the proposed planning models, we create three additional domain instances that simulate more realistic planning instances. For the Reservoir domain, we create a system with 10 reservoirs with complex reservoir formations, where a reservoir may receive water from more than one upstream reservoirs. For the HVAC domain, we simulate a building of 6 floors and 60 rooms with complex adjacencies between neighboring rooms and across building levels.  Moreover, in order to fully capture the complex mutual temperature impact of the rooms, we train a \emph{large} transition function with one hidden layer of width 256. For the Navigation domain, we reduce the feasible action range from $[-1,1]$ to $[-0.5, 0.5]$ and increase the planning horizon to 20 time steps.

In Figures~\ref{fig:scale} (a)-(c), we compare the total rewards obtained by the domain-specific rule-based policy (blue), HD-MILP-Plan (orange) and TF-Plan (red) on larger problem instances. The analysis of Figures~\ref{fig:scale} (a)-(c) shows that TF-Plan scales better compared to HD-MILP-Plan by consistently outperforming the policy, whereas HD-MILP-Plan outperforms the other two planners in two out of three domains (i.e., Reservoir and HVAC) while suffering from scalability issues in one domain (i.e., Navigation). Particularly, we find that in Navigation domain, HD-MILP-Plan sometimes does not find feasible plans with respect to the learned model and therefore returns default no-op action values leading to its poor observed performance in this domain.

In Figure~\ref{fig:time_size}, we compare the run-time performance of the proposed planning systems over all problem instances vs. problem size.  We measure the problem size as the number of variables used in the encoding, which is a function of horizon $\horizon$, number of parameters in the learned model, and number of neural network layers. We observe that as the problem size gets larger, HD-MILP-Plan takes more computational effort to solve due to the additional overhead of proving optimality, which can be mitigated by relaxing the optimality guarantee to a bound (e.g., the 20\% duality gap of MILP-S(20\%)). We also observe that as the problem sizes get larger, the effect of preprocessing bounds and strengthening constraints of the MILP-S variants pay off.  Finally, in observing that the MILP-based running times appear to increase roughly linearly in the problem size, it is critical to note that in contrast to classical discrete planning problems where running time often scales exponentially as a function of the encoding size, the scalability of MILP-based optimization can often be highly dependent on specific aspects of the problem encoding that lead to subexponential growth rates as a function of the number of parameters.  In general, it can be difficult to predict MILP optimization time as a simple function of the number of parameters being optimized.

If we examine TF-Plan in Figure~\ref{fig:time_size}, we observe that it scales gracefully as the problem size gets larger --- while these results surprisingly show that TF-Plan is fastest on the largest problem, we note that due to its highly parallel GPU-based implementation, adding more parameters does not necessarily linearly increase the running time of TF-Plan.
Furthermore, we remark that (like MILPs) continuous optimization running times can be hard to predict as a function of problem size --- small continuous problems may sometimes be very difficult to optimize while other larger continuous problems may have simpler optimization surfaces that can be optimized much more efficiently.   Together, these two reasons can help explain the somewhat surprisingly low running time of TF-Plan for the largest problem sizes.  In summary, if we also revisit Figures~\ref{fig:scale} (a)-(c), we observe that TF-Plan can provide a highly efficient alternative to HD-MILP-Plan in large-scale planning problems.

\section{Related Work}
\label{sec:related_work}

In this section, we discuss the existing automated planning literature in relation to our data-driven planners. In this work, we have focused on planning with ReLU-based DNN learned state transition models subject to the optimization of general piecewise linear reward functions over concurrent instantaneous actions with parameters that have real (continuous) domains. Many existing PDDL-based~\cite{Penna2009,coles2013hybrid,ivankovic,Bryce2015,piotrowski,scala2016interval,Scala2016-2,Cashmore2016} and RDDL-based~\cite{keller_icaps13} planners focus on models with action parameters that are strictly finite (discrete).  In order to represent more general problems, the PDDL formalism has been extended to handle action variables that have real domains (i.e., real-valued control parameters) with the focus of synthesizing partially ordered plans~\shortcite{savas2016}. While the extension of the PDDL formalism to real-valued action parameters is an important step forward towards the domain expressiveness we handle in this article, more work would be needed to show how a multilayer ReLU-based neural network transition model with piecewise nonlinearities and feedforward computations at hidden layers could be compactly and efficiently encoded in PDDL to apply such a PDDL-based planner.

In a different vein, hybrid automaton-based Domain Predictive Control can model state transition functions with linear dynamics over action parameters with real domains~\cite{lohr}. While Domain Predictive Control could technically be used to model ReLU-based state transition functions using linear state transitions, the number of mode switches required to represent each activation pattern of the learned DNN would be exponential in the number of ReLUs per time step which makes this approach ill-suited for solving the neural net transition planning problems considered in this work. 
As a further expressivity extension, hybrid planning problems with continuous (time) state transition models can be solved using a bounding technique known as \emph{flow tubes} but this approach requires the state transition function to be modeled as ordinary differential equations~\cite{li2008}, which are not well-suited to our learned neural network transition models.  Because techniques from robotics are largely specialized for the physics and geometry of those particular problems, we are not aware of techniques in robotics that can plan for arbitrary ReLU-based deep neural network state transition models as we contribute here.  As such, we conclude that our data-driven planners for learned neural net models uniquely address a complex problem related to existing literature in the field of domain-independent automated planning methods.

\section{Concluding Remarks and Future Work}
\label{sec:conclusion}

In this paper, we have tackled the question of how we can plan with expressive and accurate deep network learned transition models that are not amenable to existing solution techniques.  We started by focusing on how to learn accurate learned transition functions by using densely-connected deep neural networks trained with a reweighted mean square error (MSE) loss. We then leveraged the insight that ReLU-based deep networks offer strong learning performance and permit a 
direct compilation of the neural network transition model to a Mixed-Integer Linear Program (MILP) encoding in a planner we called Hybrid Deep MILP Planner (HD-MILP-Plan). To further enhance planning efficiency, we strengthened the linear relaxation of the base MILP encoding. Finally, as a more efficient but not provably optimal alternative to MILP-based optimization, we proposed the end-to-end gradient optimization-based Tensorflow Planner (TF-Plan) that encodes the problem in the form of a Recurrent Neural Network (RNN) and directly optimizes plans via backpropagation.

We evaluated run-time performance and solution quality of the plans generated by both proposed planners over multiple problem instances from three diverse continuous state and action planning domains. It would be hard to definitively characterize the general comparative performance behavior of HD-MILP-Plan vs.\ TF-Plan for arbitrary problems; however, on these particular domains, we have shown that HD-MILP-Plan can find optimal plans with respect to the learned models, and TF-Plan can approximate the optimal plans with little computational cost.  We have shown that the plans generated by both HD-MILP-Plan and TF-Plan yield better solution qualities compared to strong domain-specific human-designed policies and that TF-Plan performance generally improves with the number of optimization epochs. Also, we have shown that our strengthening constraints improved the solution quality and the run-time performance of HD-MILP-Plan as problem instances got larger. Finally, we have shown that TF-Plan can handle large-scale planning problems with very little relative computational cost compared to the MILP-based optimization approach of HD-MILP Plan and its variants defined in Section~\ref{MILP}.


In terms of future work, while TF-Plan showed strong performance and scalability for large problems evaluated in this article, it does have two important weaknesses compared to HD-MILP-Plan that provide avenues for additional research.  First, while HD-MILP-Plan in principle has no problem working with discrete actions --- learning with discrete actions requires no special encodings over that previously given and MILP solvers directly support discrete variables --- the only way to currently handle discrete actions in TF-Plan would be through a continuous relaxation.  Second, while goal and state constraints are straightforward to handle in the HD-MILP-Plan framework since these are literally just additional constraints in the encoding, Tensorflow is generally not intended for constrained optimization.  To this end, future work should consider how TF-Plan can be effectively extended to handle discrete actions and general constraints.  Finally, we remark that this article did not address discrete states, with a key caveat simply being that effective learning of deterministic deep neural networks with discrete output nodes remains an open area of research investigation in deep neural networks.  That said, we note that some recent work on planning with learned binarized neural network (BNN) transition models using MILP, SAT and pseudo-Boolean encodings~\shortcite{say:ijcai18a,Say2018b,Say2020b} does pose one direction for future research with discrete states in combination with the contributions of this work focused on continuous states.

In conclusion, both HD-MILP-Plan and TF-Plan represent \emph{a new class of data-driven planning methods} that can accurately learn complex state transitions of high-dimensional nonlinear continuous state and action planning domains, and provide high-quality plans with respect to these  learned models.  Further, we believe that this work paves the way for future research that extends HD-MILP-Plan and TF-Plan to learning and optimizing in transition models with discrete states as well as further extensions to TF-Plan for handling discrete actions and general constraints in a highly general and highly scalable Tensorflow-based planner.

\section*{Acknowledgements}


Preliminary versions of this article appeared in conference papers~\shortcite{Say2017} and~\shortcite{Wu2017}. Both Ga Wu and Buser Say contributed equally to this work. For Buser Say, the majority of the work was done while he was at the University of Toronto and affiliated with the Vector Institute, and only the final revision of the article was done while the he was at the Monash University. This work was supported by an NSERC Discovery Grant and an Ontario Early Researcher Award.

\bibliography{sample}
\bibliographystyle{theapa}


\newpage

\appendix
\section{Post-learning Neural Network Weight Modification}
\label{app:normalization}
 



As outlined in Section~\ref{sec:normalization}, we standardize input $\bar{\myvec{X}} = \bar{\myvec{S}}_n{}^\frown \bar{\myvec{A}}_n$ before feeding it to the neural network. Let $\bar{\myvec{\mu}}$ and $\bar{\myvec{\sigma}}$ respectively denote the vector of means and standard deviations for $\bar{\myvec{X}}$ empirically calculated from the training data.  Then the normalized inputs $\hat{\bar{\myvec{X}}}$ used to train the deep neural network are defined as 
\[\hat{\bar{\myvec{X}}} = (\bar{\myvec{X}}-\bar{\myvec{\mu}})\cdot\bar{\myvec{\sigma}}^{-1} \, . \]

Now, let $\myvec{W}$ denote the vector of weights connected to a hidden unit of a neural network, let $b$ indicate the bias term for this unit, and let $\bar{z}$ denote the value of the hidden unit before the nonlinear activation is applied: 
\[\bar{z} = \hat{\bar{\myvec{X}}}^T\myvec{W} + b \, .\]

Our objective now is to express $z$ in terms of $\bar{\myvec{X}}$ and determine a modified weight vector $\myvec{W}'$ and bias $b'$ that can be used to replace $\myvec{W}$ and $b$ in order to allow the network to accept unnormalized inputs.  This derivation requires some simple algebraic manipulation, which we show below:
\begin{equation}
\begin{split}
\bar{z} &= \hat{\bar{\myvec{X}}}^T\myvec{W} + b\\
&= ((\bar{\myvec{X}}-\bar{\myvec{\mu}})\cdot\bar{\myvec{\sigma}}^{-1})^T\myvec{W} + b\\
&= (\bar{\myvec{X}}-\bar{\myvec{\mu}})^T(\myvec{W}\cdot\bar{\myvec{\sigma}}^{-1}) + b\\
&= \bar{\myvec{X}}^T\underbrace{(\myvec{W}\cdot\bar{\myvec{\sigma}}^{-1})}_{\myvec{W}'} + \underbrace{b - \bar{\myvec{\mu}}^T(\myvec{W}\cdot\bar{\myvec{\sigma}}^{-1})}_{b'} \\
\end{split}
\end{equation}
We remark that this transformation need only apply to the weights and biases of units that connect directly to the input layer.

\section{RDDL Domain Descriptions}
\label{app:domain_descriptions}

In this section, we list the RDDL domain and instance files that we experimented with in this paper.
\subsection*{Reservoir}
\subsubsection*{Domain File}
\begin{verbatim}
domain Reservoir_Problem{

	requirements = { 
		reward-deterministic 
	};

	types {
		id: object;
	};
	
	pvariables {
	
		// Constant
		MAXCAP(id): { non-fluent, real, default = 100.0 };
		HIGH_BOUND(id): { non-fluent, real, default = 80.0 };
		LOW_BOUND(id): { non-fluent, real, default = 20.0 };
		RAIN(id): { non-fluent ,real, default = 5.0 };
		DOWNSTREAM(id,id): {non-fluent ,bool, default = false };
		DOWNTOSEA(id): {non-fluent, bool, default = false };
		BIGGESTMAXCAP: {non-fluent, real, default = 1000};

		//Interm
		vaporated(id): {interm-fluent, real};
		
		//State
		rlevel(id): {state-fluent, real, default = 50.0 };
		
		//Action
		flow(id): { action-fluent, real, default = 0.0 };
	};
	
	cpfs {
		vaporated(?r) = (1.0/2.0)*sin[rlevel(?r)/BIGGESTMAXCAP]*rlevel(?r);
		rlevel'(?r) = rlevel(?r) + RAIN(?r)- vaporated(?r) - flow(?r) 
		              + sum_{?r2: id}[DOWNSTREAM(?r2,?r)*flow(?r2)];

	};
	
	reward = sum_{?r: id} [if (rlevel'(?r)>=LOW_BOUND(?r) ^ (rlevel'(?r)<=HIGH_BOUND(?r)))
			                       then 0
			                       else if (rlevel'(?r)<=LOW_BOUND(?r))
			                               then (-5)*(LOW_BOUND(?r)-rlevel'(?r))
			                               else (-100)*(rlevel'(?r)-HIGH_BOUND(?r))]
	         +sum_{?r2:id}[abs[((HIGH_BOUND(?r2)+LOW_BOUND(?r2))/2.0)-rlevel'(?r2)]*(-0.1)];
								
	state-action-constraints {
	
		forall_{?r:id} flow(?r)<=rlevel(?r);
		forall_{?r:id} rlevel(?r)<=MAXCAP(?r);
		forall_{?r:id} flow(?r)>=0;
	};
}
\end{verbatim}
\subsubsection*{Instance Files}
Reservoir 3
\begin{verbatim}
non-fluents Reservoir_non {
    domain = Reservoir_Problem;
    objects{
        id: {t1,t2,t3};
    };
    non-fluents {
        RAIN(t1) = 5.0;RAIN(t2) = 10.0;RAIN(t3) = 20.0;
        MAXCAP(t2) = 200.0;LOW_BOUND(t2) = 30.0;HIGH_BOUND(t2) = 180.0;
        MAXCAP(t3) = 400.0;LOW_BOUND(t3) = 40.0;HIGH_BOUND(t3) = 380.0;
        DOWNSTREAM(t1,t2);DOWNSTREAM(t2,t3);DOWNTOSEA(t3);
    };
}

instance is1{
    domain = Reservoir_Problem;
    non-fluents = Reservoir_non;
    init-state{
        rlevel(t1) = 75.0;
    };
    max-nondef-actions = 3;
    horizon = 10;
    discount = 1.0;
}
\end{verbatim}
Reservoir 4
\begin{verbatim}
non-fluents Reservoir_non {
    domain = Reservoir_Problem;
    objects{
        id: {t1,t2,t3,t4};
    };
    non-fluents {
        RAIN(t1) = 5.0;RAIN(t2) = 10.0;RAIN(t3) = 20.0;RAIN(t4) = 30.0;
        MAXCAP(t2) = 200.0;LOW_BOUND(t2) = 30.0;HIGH_BOUND(t2) = 180.0;
        MAXCAP(t3) = 400.0;LOW_BOUND(t3) = 40.0;HIGH_BOUND(t3) = 380.0;
        MAXCAP(t4) = 500.0;LOW_BOUND(t4) = 60.0;HIGH_BOUND(t4) = 480.0;
        DOWNSTREAM(t1,t2);DOWNSTREAM(t2,t3);DOWNSTREAM(t3,t4);DOWNTOSEA(t4);
    };
}

instance is1{
    domain = Reservoir_Problem;
    non-fluents = Reservoir_non;
    init-state{
        rlevel(t1) = 75.0;
    };
    max-nondef-actions = 4;
    horizon = 10;
    discount = 1.0;
}
\end{verbatim}
Reservoir 10
\begin{verbatim}
non-fluents Reservoir_non {
	domain = Reservoir_Problem;
	objects{
		id: {t1,t2,t3,t4,t5,t6,t7,t8,t9,t10};
	};
	non-fluents {
		RAIN(t1) = 15.0;RAIN(t2) = 10.0;RAIN(t3) = 20.0;RAIN(t4) = 30.0;RAIN(t5) = 20.0;
		RAIN(t6) = 10.0;RAIN(t7) = 35.0;RAIN(t8) = 15.0;RAIN(t9) = 25.0;RAIN(t10) = 20.0;
		MAXCAP(t2) = 200.0;LOW_BOUND(t2) = 30.0;HIGH_BOUND(t2) = 180.0;
		MAXCAP(t3) = 400.0;LOW_BOUND(t3) = 40.0;HIGH_BOUND(t3) = 380.0;
		MAXCAP(t4) = 500.0;LOW_BOUND(t4) = 60.0;HIGH_BOUND(t4) = 480.0;
		MAXCAP(t5) = 750.0;LOW_BOUND(t5) = 20.0;HIGH_BOUND(t5) = 630.0;
		MAXCAP(t6) = 300.0;LOW_BOUND(t6) = 30.0;HIGH_BOUND(t6) = 250.0;
		MAXCAP(t7) = 300.0;LOW_BOUND(t7) = 10.0;HIGH_BOUND(t7) = 180.0;
		MAXCAP(t8) = 300.0;LOW_BOUND(t8) = 40.0;HIGH_BOUND(t8) = 240.0;
		MAXCAP(t9) = 400.0;LOW_BOUND(t9) = 40.0;HIGH_BOUND(t9) = 340.0;
		MAXCAP(t10) = 800.0;LOW_BOUND(t10) = 20.0;HIGH_BOUND(t10) = 650.0;
		DOWNSTREAM(t1,t2);DOWNSTREAM(t2,t3);DOWNSTREAM(t3,t4);DOWNSTREAM(t4,t5);
		DOWNSTREAM(t6,t7);DOWNSTREAM(t7,t8);DOWNSTREAM(t8,t5);
		DOWNSTREAM(t5,t6);DOWNSTREAM(t6,t10);
		DOWNSTREAM(t5,t9);DOWNSTREAM(t9,t10);
		DOWNTOSEA(t10);
	};
}

instance is1{
	domain = Reservoir_Problem;
	non-fluents = Reservoir_non;
	init-state{
		rlevel(t1) = 175.0;
	};
	max-nondef-actions = 10;
	horizon = 10;
	discount = 1.0;
}
\end{verbatim}
\subsection*{HVAC}
\subsubsection*{Domain File}
\begin{verbatim}
domain hvac_vav_fix{
	  types {
		  space : object;
  	};

  	pvariables {
  		//Constants
  		ADJ(space, space)    : { non-fluent, bool, default = false };
  		ADJ_OUTSIDE(space)		: { non-fluent, bool, default = false };
  		ADJ_HALL(space)			: { non-fluent, bool, default = false };
  		R_OUTSIDE(space)		: { non-fluent, real, default = 4};
    	R_HALL(space)			: { non-fluent, real, default = 2};
  		R_WALL(space, space) : { non-fluent, real, default = 1.5 };
  		IS_ROOM(space)		 : { non-fluent, bool, default = false };
  		CAP(space) 			 : { non-fluent, real, default = 80 }; 
  		CAP_AIR 			 : { non-fluent, real, default = 1.006 }; 
  		COST_AIR 			 : { non-fluent, real, default = 1 };	
  		TIME_DELTA 			 : { non-fluent, real, default = 1 }; 
  		TEMP_AIR 			 : { non-fluent, real, default = 40 }; 
  		TEMP_UP(space)		 : { non-fluent, real, default = 23.5 }; 
  		TEMP_LOW(space)		 : { non-fluent, real, default = 20.0 }; 
  		TEMP_OUTSIDE(space)		: { non-fluent, real, default = 6.0 }; 
  		TEMP_HALL(space)		: { non-fluent, real, default = 10.0 }; 
  		PENALTY 			 : { non-fluent, real, default = 20000 }; 
		AIR_MAX(space)		 : { non-fluent, real, default = 10.0 }; 
  		TEMP(space) 	     : { state-fluent, real, default = 10.0 };	
  		AIR(space)		     : { action-fluent, real, default = 0.0 }; 
  	};

  	cpfs {
  		//State
  		TEMP'(?s) = TEMP(?s) + TIME_DELTA/CAP(?s) * 
 			 (AIR(?s) * CAP_AIR * (TEMP_AIR - TEMP(?s)) * IS_ROOM(?s) 
  			+ sum_{?p : space} ((ADJ(?s, ?p) | ADJ(?p, ?s)) * (TEMP(?p) - TEMP(?s)) / R_WALL(?s, ?p))
  			+ ADJ_OUTSIDE(?s)*(TEMP_OUTSIDE(?s) - TEMP(?s))/ R_OUTSIDE(?s) 
 			+ ADJ_HALL(?s)*(TEMP_HALL(?s)-TEMP(?s))/R_HALL(?s));
		};
		

  	reward = - (sum_{?s : space} IS_ROOM(?s)*(AIR(?s) * COST_AIR
             + ((TEMP(?s) < TEMP_LOW(?s)) | (TEMP(?s) > TEMP_UP(?s))) * PENALTY) 
             + 10.0*abs[((TEMP_UP(?s) + TEMP_LOW(?s))/2.0) - TEMP(?s)]);

    action-preconditions{
			forall_{?s : space} [ AIR(?s) >= 0 ];
			forall_{?s : space} [ AIR(?s) <= AIR_MAX(?s)];
		};
}
\end{verbatim}
\subsubsection*{Instance Files}
HVAC 3 Rooms
\begin{verbatim}
non-fluents nf_hvac_vav_fix{
    domain = hvac_vav_fix;

    objects{
        space : { r1, r2, r3}; 
    };

    non-fluents {
        //Define rooms
        IS_ROOM(r1) = true;IS_ROOM(r2) = true;IS_ROOM(r3) = true;
        
        //Define the adjacency
        ADJ(r1, r2) = true;ADJ(r1, r3) = true;ADJ(r2, r3) = true;
        
        ADJ_OUTSIDE(r1) = true;ADJ_OUTSIDE(r2) = true;
        ADJ_HALL(r1) = true;ADJ_HALL(r3) = true;
    };
}

instance inst_hvac_vav_fix{
    domain = hvac_vav_fix;
    non-fluents = nf_hvac_vav_fix;

    horizon = 20;
    discount = 1.0;
}
\end{verbatim}
HVAC 6 Rooms
\begin{verbatim}
non-fluents nf_hvac_vav_fix{
    domain = hvac_vav_fix;

    objects{
        space : { r1, r2, r3, r4, r5, r6 }; 
    };

    non-fluents {
        //Define rooms
        IS_ROOM(r1) = true;IS_ROOM(r2) = true;IS_ROOM(r3) = true;
        IS_ROOM(r4) = true;IS_ROOM(r5) = true;IS_ROOM(r6) = true;
        
        //Define the adjacency
        ADJ(r1, r2) = true;ADJ(r1, r4) = true;ADJ(r2, r3) = true;
        ADJ(r2, r5) = true;ADJ(r3, r6) = true;ADJ(r4, r5) = true;
        ADJ(r5, r6) = true;
        
        ADJ_OUTSIDE(r1) = true;ADJ_OUTSIDE(r3) = true;
        ADJ_OUTSIDE(r4) = true;ADJ_OUTSIDE(r6) = true;
        ADJ_HALL(r1) = true;ADJ_HALL(r2) = true;ADJ_HALL(r3) = true;
        ADJ_HALL(r4) = true;ADJ_HALL(r5) = true;ADJ_HALL(r6) = true;
    };
}

instance inst_hvac_vav_fix{
    domain = hvac_vav_fix;
    non-fluents = nf_hvac_vav_fix;

    horizon = 20;
    discount = 1.0;
}
\end{verbatim}
HVAC 60 Rooms
\begin{verbatim}
non-fluents nf_hvac_vav_fix{
    domain = hvac_vav_fix;

    objects{
        space : { r101, r102, r103, r104, r105, r106, r107, r108, r109, r110, r111, r112,
              r201, r202, r203, r204, r205, r206, r207, r208, r209, r210, r211, r212,
              r301, r302, r303, r304, r305, r306, r307, r308, r309, r310, r311, r312,
              r401, r402, r403, r404, r405, r406, r407, r408, r409, r410, r411, r412,
              r501, r502, r503, r504, r505, r506, r507, r508, r509, r510, r511, r512
             }; //Three rooms, one hallway, and the outside world
    };

    non-fluents {
        //Define rooms
        //Level1
        IS_ROOM(r101) = true;IS_ROOM(r102) = true;IS_ROOM(r103) = true;IS_ROOM(r104) = true;
        IS_ROOM(r105) = true;IS_ROOM(r106) = true;IS_ROOM(r107) = true;IS_ROOM(r108) = true;
        IS_ROOM(r109) = true;IS_ROOM(r110) = true;IS_ROOM(r111) = true;IS_ROOM(r112) = true;
        //Level2
        IS_ROOM(r201) = true;IS_ROOM(r202) = true;IS_ROOM(r203) = true;IS_ROOM(r204) = true;
        IS_ROOM(r205) = true;IS_ROOM(r206) = true;IS_ROOM(r207) = true;IS_ROOM(r208) = true;
        IS_ROOM(r209) = true;IS_ROOM(r210) = true;IS_ROOM(r211) = true;IS_ROOM(r212) = true;
        //Level3
        IS_ROOM(r301) = true;IS_ROOM(r302) = true;IS_ROOM(r303) = true;IS_ROOM(r304) = true;
        IS_ROOM(r305) = true;IS_ROOM(r306) = true;IS_ROOM(r307) = true;IS_ROOM(r308) = true;
        IS_ROOM(r309) = true;IS_ROOM(r310) = true;IS_ROOM(r311) = true;IS_ROOM(r312) = true;
        //Level4
        IS_ROOM(r401) = true;IS_ROOM(r402) = true;IS_ROOM(r403) = true;IS_ROOM(r404) = true;
        IS_ROOM(r405) = true;IS_ROOM(r406) = true;IS_ROOM(r407) = true;IS_ROOM(r408) = true;
        IS_ROOM(r409) = true;IS_ROOM(r410) = true;IS_ROOM(r411) = true;IS_ROOM(r412) = true;
        //Level5
        IS_ROOM(r501) = true;IS_ROOM(r502) = true;IS_ROOM(r503) = true;IS_ROOM(r504) = true;
        IS_ROOM(r505) = true;IS_ROOM(r506) = true;IS_ROOM(r507) = true;IS_ROOM(r508) = true;
        IS_ROOM(r509) = true;IS_ROOM(r510) = true;IS_ROOM(r511) = true;IS_ROOM(r512) = true;


        //Define the adjacency
        //Level1
        ADJ(r101, r102) = true;ADJ(r102, r103) = true;ADJ(r103, r104) = true;
        ADJ(r104, r105) = true;ADJ(r106, r107) = true;ADJ(r107, r108) = true;
        ADJ(r107, r109) = true;ADJ(r108, r109) = true;ADJ(r110, r111) = true;
        ADJ(r111, r112) = true;
        //Level2
        ADJ(r201, r202) = true;ADJ(r202, r203) = true;ADJ(r203, r204) = true;
        ADJ(r204, r205) = true;ADJ(r206, r207) = true;ADJ(r207, r208) = true;
        ADJ(r207, r209) = true;ADJ(r208, r209) = true;ADJ(r210, r211) = true;
        ADJ(r211, r212) = true;
        //Level3
        ADJ(r301, r302) = true;ADJ(r302, r303) = true;ADJ(r303, r304) = true;
        ADJ(r304, r305) = true;ADJ(r306, r307) = true;ADJ(r307, r308) = true;
        ADJ(r307, r309) = true;ADJ(r308, r309) = true;ADJ(r310, r311) = true;
        ADJ(r311, r312) = true;
        //Level4
        ADJ(r401, r402) = true;ADJ(r402, r403) = true;ADJ(r403, r404) = true;
        ADJ(r404, r405) = true;ADJ(r406, r407) = true;ADJ(r407, r408) = true;
        ADJ(r407, r409) = true;ADJ(r408, r409) = true;ADJ(r410, r411) = true;
        ADJ(r411, r412) = true;
        //Level5
        ADJ(r501, r502) = true;ADJ(r502, r503) = true;ADJ(r503, r504) = true;
        ADJ(r504, r505) = true;ADJ(r506, r507) = true;ADJ(r507, r508) = true;
        ADJ(r507, r509) = true;ADJ(r508, r509) = true;ADJ(r510, r511) = true;
        ADJ(r511, r512) = true;
        //InterLevel 1-2
        ADJ(r101, r201) = true;ADJ(r102, r202) = true;ADJ(r103, r203) = true;
        ADJ(r104, r204) = true;ADJ(r105, r205) = true;ADJ(r106, r206) = true;
        ADJ(r107, r207) = true;ADJ(r108, r208) = true;ADJ(r109, r209) = true;
        ADJ(r110, r210) = true;ADJ(r111, r211) = true;ADJ(r112, r212) = true;
        //InterLevel 2-3
        ADJ(r201, r301) = true;ADJ(r202, r302) = true;ADJ(r203, r303) = true;
        ADJ(r204, r304) = true;ADJ(r205, r305) = true;ADJ(r206, r306) = true;
        ADJ(r207, r307) = true;ADJ(r208, r308) = true;ADJ(r209, r309) = true;
        ADJ(r210, r310) = true;ADJ(r211, r311) = true;ADJ(r212, r312) = true;
        //InterLevel 3-4
        ADJ(r301, r401) = true;ADJ(r302, r402) = true;ADJ(r303, r403) = true;
        ADJ(r304, r404) = true;ADJ(r305, r405) = true;ADJ(r306, r406) = true;
        ADJ(r307, r407) = true;ADJ(r308, r408) = true;ADJ(r309, r409) = true;
        ADJ(r310, r410) = true;ADJ(r311, r411) = true;ADJ(r312, r412) = true;
        //InterLevel 4-5
        ADJ(r401, r501) = true;ADJ(r402, r502) = true;ADJ(r403, r503) = true;
        ADJ(r404, r504) = true;ADJ(r405, r505) = true;ADJ(r406, r506) = true;
        ADJ(r407, r507) = true;ADJ(r408, r508) = true;ADJ(r409, r509) = true;
        ADJ(r410, r510) = true;ADJ(r411, r511) = true;ADJ(r412, r512) = true;

        
        //Outside
        //Level1
        ADJ_OUTSIDE(r101) = true;ADJ_OUTSIDE(r102) = true;ADJ_OUTSIDE(r103) = true;
        ADJ_OUTSIDE(r104) = true;ADJ_OUTSIDE(r105) = true;ADJ_OUTSIDE(r106) = true;
        ADJ_OUTSIDE(r108) = true;ADJ_OUTSIDE(r110) = true;ADJ_OUTSIDE(r111) = true;
        ADJ_OUTSIDE(r112) = true;
        //Level2
        ADJ_OUTSIDE(r201) = true;ADJ_OUTSIDE(r202) = true;ADJ_OUTSIDE(r203) = true;
        ADJ_OUTSIDE(r204) = true;ADJ_OUTSIDE(r205) = true;ADJ_OUTSIDE(r206) = true;
        ADJ_OUTSIDE(r208) = true;ADJ_OUTSIDE(r210) = true;ADJ_OUTSIDE(r211) = true;
        ADJ_OUTSIDE(r212) = true;
        //Level3
        ADJ_OUTSIDE(r301) = true;ADJ_OUTSIDE(r302) = true;ADJ_OUTSIDE(r303) = true;
        ADJ_OUTSIDE(r304) = true;ADJ_OUTSIDE(r305) = true;ADJ_OUTSIDE(r306) = true;
        ADJ_OUTSIDE(r308) = true;ADJ_OUTSIDE(r310) = true;ADJ_OUTSIDE(r311) = true;
        ADJ_OUTSIDE(r312) = true;
        //Level4
        ADJ_OUTSIDE(r401) = true;ADJ_OUTSIDE(r402) = true;ADJ_OUTSIDE(r403) = true;
        ADJ_OUTSIDE(r404) = true;ADJ_OUTSIDE(r405) = true;ADJ_OUTSIDE(r406) = true;
        ADJ_OUTSIDE(r408) = true;ADJ_OUTSIDE(r410) = true;ADJ_OUTSIDE(r411) = true;
        ADJ_OUTSIDE(r412) = true;
        //Level5
        ADJ_OUTSIDE(r501) = true;ADJ_OUTSIDE(r502) = true;ADJ_OUTSIDE(r503) = true;
        ADJ_OUTSIDE(r504) = true;ADJ_OUTSIDE(r505) = true;ADJ_OUTSIDE(r506) = true;
        ADJ_OUTSIDE(r508) = true;ADJ_OUTSIDE(r510) = true;ADJ_OUTSIDE(r511) = true;
        ADJ_OUTSIDE(r512) = true;

        //Hallway
        //Level1
        ADJ_HALL(r101) = true;ADJ_HALL(r102) = true;ADJ_HALL(r103) = true;
        ADJ_HALL(r106) = true;ADJ_HALL(r107) = true;ADJ_HALL(r109) = true;
        ADJ_HALL(r110) = true;
        //Level2
        ADJ_HALL(r201) = true;ADJ_HALL(r202) = true;ADJ_HALL(r203) = true;
        ADJ_HALL(r206) = true;ADJ_HALL(r207) = true;ADJ_HALL(r209) = true;
        ADJ_HALL(r210) = true;
        //Level3
        ADJ_HALL(r301) = true;ADJ_HALL(r302) = true;ADJ_HALL(r303) = true;
        ADJ_HALL(r306) = true;ADJ_HALL(r307) = true;ADJ_HALL(r309) = true;
        ADJ_HALL(r310) = true;
        //Level4
        ADJ_HALL(r401) = true;ADJ_HALL(r402) = true;ADJ_HALL(r403) = true;
        ADJ_HALL(r406) = true;ADJ_HALL(r407) = true;ADJ_HALL(r409) = true;
        ADJ_HALL(r410) = true;
        //Level5
        ADJ_HALL(r501) = true;ADJ_HALL(r502) = true;ADJ_HALL(r503) = true;
        ADJ_HALL(r506) = true;ADJ_HALL(r507) = true;ADJ_HALL(r509) = true;
        ADJ_HALL(r510) = true;
    };
}

instance inst_hvac_vav_fix{
    domain = hvac_vav_fix;
    non-fluents = nf_hvac_vav_fix;
    //init-state{
    //};
    max-nondef-actions = 60;
    horizon = 12;
    discount = 1.0;
}
\end{verbatim}
\subsection*{Navigation}
\subsubsection*{Domain File}
\begin{verbatim}
domain Navigation_Problem{

    requirements = { 
        reward-deterministic 
    };

    types {
        dim: object;
    };
    
    pvariables {
    
        // Constant
        MINMAZEBOUND(dim): { non-fluent, real, default = -4.0 }; //-5.0 for 10x10 instance
        MAXMAZEBOUND(dim): { non-fluent, real, default = 4.0 }; //5.0 for 10x10 instance
        MINACTIONBOUND(dim): { non-fluent, real, default = -1.0 }; //-0.5 for large scale instance
        MAXACTIONBOUND(dim): { non-fluent, real, default = 1.0 }; //0.5 for large scale instance
        GOAL(dim): { non-fluent, real, default = 3.0 };
        PENALTY: {non-fluent, real, default = 1000000.0 };
        CENTER(dim): {non-fluent, real, default = 0.0};
        
        // Interm        
        distance: {interm-fluent,real,level=1 };
        scalefactor: {interm-fluent,real,level=2 };
        proposedLoc(dim):{interm-fluent, real, level=3};
 
        //State
        location(dim): {state-fluent, real, default = -4.0 }; //-5.0 for 10x10 instance
                
        //Action
        move(dim): { action-fluent, real, default = 0.0 };
    };
    
    cpfs {

        distance = sqrt[sum_{?l:dim}[pow[(location(?l)-CENTER(?l)),2]]];
        scalefactor = 2.0/(1.0+exp[-2*distance])-0.99;
        proposedLoc(?l) = location(?l) + move(?l)*scalefactor;
        location'(?l)= if(proposedLoc(?l)<=MAXMAZEBOUND(?l) ^ proposedLoc(?l)>=MINMAZEBOUND(?l)) 
                       then proposedLoc(?l) 
                       else (if(proposedLoc(?l)>MAXMAZEBOUND(?l)) 
                             then MAXMAZEBOUND(?l) else MINMAZEBOUND(?l));


    };
    
    reward = - sum_{?l: dim}[abs[GOAL(?l) - location(?l)]];
                                
    state-action-constraints {
        forall_{?l:dim} move(?l)<=MAXACTIONBOUND(?l);
        forall_{?l:dim} move(?l)>=MINACTIONBOUND(?l);
        forall_{?l:dim} location(?l)<=MAXMAZEBOUND(?l);
        forall_{?l:dim} location(?l)>=MINMAZEBOUND(?l);
    };
}
\end{verbatim}
\subsubsection*{Instance Files}
Navigation 8 by 8 instance
\begin{verbatim}
non-fluents Navigation_non {
    domain = Navigation_Problem;
    objects{
        dim: {x,y};
    };
    non-fluents {
        MINMAZEBOUND(x) = -4.0;
    };
}

instance is1{
    domain = Navigation_Problem;
    non-fluents = Navigation_non;
    init-state{
        location(x) = -4.0;location(y) = -4.0;
    };
    max-nondef-actions = 2;
    horizon = 10;
    discount = 1.0;
}
\end{verbatim}
Navigation 10 by 10 instance
\begin{verbatim}
non-fluents Navigation_non {
    domain = Navigation_Problem;
    objects{
        dim: {x,y};
    };
    non-fluents {
        MINMAZEBOUND(x) = -5.0;
    };
}

instance is1{
    domain = Navigation_Problem;
    non-fluents = Navigation_non;
    init-state{
        location(x) = -5.0;location(y) = -5.0;
    };
    max-nondef-actions = 2;
    horizon = 10; //20 for large scale instance
    discount = 1.0;
}
\end{verbatim}

\end{document}